\documentclass[10pt,twocolumn,letterpaper]{article}
\usepackage[pagenumbers]{cvpr}
\usepackage[T1]{fontenc}
\usepackage{float}
\usepackage{booktabs}
\usepackage[table]{xcolor}
\usepackage{graphicx}
\usepackage{tabularx}
\usepackage{graphicx}
\usepackage{amsmath,amssymb,booktabs,multirow}
\usepackage{xcolor}
\usepackage{xspace}

\usepackage{cuted}

\makeatletter
\@ifpackageloaded{lineno}{%
  \IfFormatAtLeastTF{2025-06-01}{%
    \RemoveFromHook{build/column/before}[lineno]
    \AddToHook{build/column/after}[cvpr-lineno-columns]{%
      \setbox\@outputbox\vbox{%
        \boxmaxdepth\@maxdepth
        \protected@write\@auxout{}{%
          \string\def\string\@LN@column{\if@firstcolumn1\else2\fi}%
        }%
        \box\@outputbox
      }%
    }%
  }{}%
}{}%
\makeatother

\definecolor{cvprblue}{rgb}{0.21,0.49,0.74}
\usepackage[pagebackref,breaklinks,colorlinks,allcolors=cvprblue]{hyperref}
\hypersetup{pdftitle={PanoTrek: Generating Explorable 3D Worlds via
Camera-Controlled Panoramic Video},pdfauthor={Jiafu He, Chunyu Song, Dingxi Yang, Wenqi Guo, Zhijin Qin, Zhongqing Han}}
\def\confName{CVPR}
\def\confYear{2027}
\title{PanoTrek: Generating Explorable 3D Worlds via\\ Camera-Controlled Panoramic Video}
\author{
{\fontsize{12}{14}\selectfont\bfseries Jiafu He$^{1,*}$\hspace{1.25em}Chunyu Song$^{2,*}$\hspace{1.25em}Dingxi Yang$^{1}$}\\
{\fontsize{12}{14}\selectfont\bfseries Wenqi Guo$^{1}$\hspace{1.25em}Zhijin Qin$^{1,\dagger}$\hspace{1.25em}Zhongqing Han$^{2,\ddagger}$}\\
{\normalsize $^{1}$Tsinghua University\quad $^{2}$Anker Innovation}\\[-1pt]
{\normalsize\href{mailto:hjf25@mails.tsinghua.edu.cn}{\textcolor{black}{hjf25@mails.tsinghua.edu.cn}}\quad
\href{mailto:qinzhijin@tsinghua.edu.cn}{\textcolor{black}{qinzhijin@tsinghua.edu.cn}}}\\[-1pt]
{\normalsize $^{*}$Equal contribution.\quad $^{\dagger}$Corresponding author.\quad $^{\ddagger}$Project leader.}
}
\makeatletter
\def\@maketitle{%
  \newpage\null\vskip .25in
  \begin{center}
    {\fontsize{14.4}{16}\selectfont\bfseries\@title\par}
    \vskip 10pt
    {\fontsize{10}{11}\selectfont
     \begin{tabular}[t]{@{}c@{}}\@author\end{tabular}\par}
  \end{center}
}
\makeatother
\begin{document}
\maketitle

\begin{strip}
    \centering
    \includegraphics[width=0.94\textwidth]{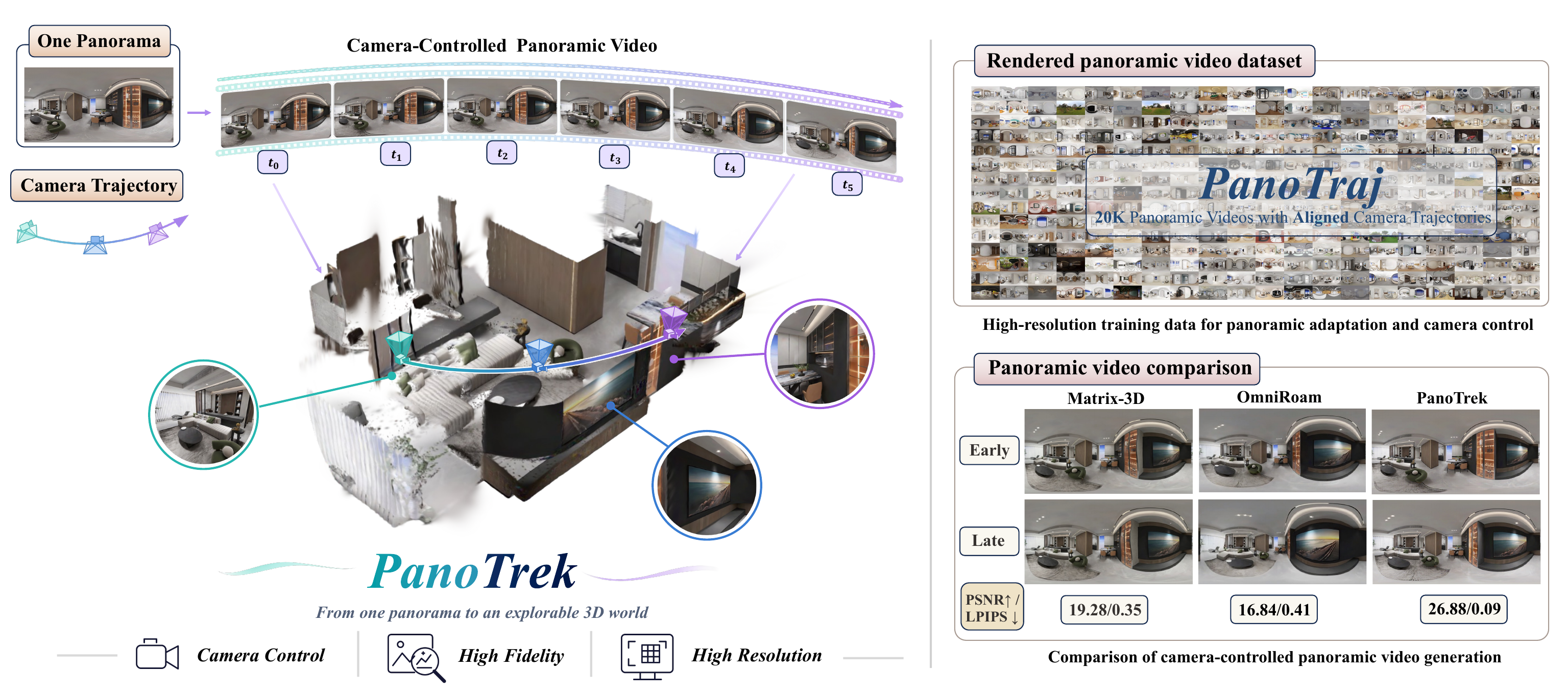}
    \captionsetup{hypcap=false}
    \captionof{figure}{\textbf{Overview of PanoTrek.}
    Given a panorama and camera trajectories, PanoTrek generates panoramic
    observations for 3DGS reconstruction and novel-view rendering.
    PanoTraj supports panoramic adaptation and camera-control training with
    high-resolution sequences and calibrated camera trajectories.
    Compared with the baselines, PanoTrek better preserves scene structure
    and fine appearance details as the camera moves.}

    \label{fig:gaussworld_overview}

    \captionsetup{hypcap=true}
\end{strip}

\begin{abstract}
Creating high-fidelity, explorable 3D worlds from a single panorama offers a promising route to immersive content creation.
However, a panorama captures the scene from one position, while reconstruction requires detailed, camera-aligned, geometrically consistent views of unobserved regions.
Learning to generate such views is constrained by the scarcity of high-resolution, high-quality panoramic videos spanning diverse scenes.
We present \textbf{PanoTrek}, a framework that generates panoramic videos as multi-view observations for explicit 3D reconstruction.
We construct \textbf{PanoTraj}, a rendered panoramic video dataset spanning indoor and outdoor environments, with calibrated camera trajectories and resolutions up to $2880 \times 1440$.
Using these data, our two-stage training adapts pretrained video priors to panoramic video generation, then learns camera control from coarse geometric renderings and their confidence maps.
To maintain consistency across video chunks, \textbf{3DCache} retains and reprojects previously generated scene content to guide subsequent synthesis.
The generated observations and prescribed camera poses are then used to reconstruct a 3D Gaussian scene for novel-view rendering.
Experiments demonstrate state-of-the-art panoramic video fidelity and camera controllability, with reconstructed scenes exhibiting substantially higher rendering quality than those of competing methods.
\end{abstract}
\section{Introduction}
\label{sec:intro}

Panoramic images capture the surroundings in all viewing directions, making them a natural starting point for immersive scene creation.
However, a panorama only observes the scene from one camera position and leaves occluded surfaces unknown.
Extending this observation to an explorable 3D scene requires synthesizing what becomes visible as the camera moves.
Recent camera-controlled panoramic video models provide such observations along prescribed trajectories~\cite{matrix3d2025,panoworldx2025,panoworld_real2026,omniroam2026}.
For reconstruction, however, visual plausibility alone is insufficient: the generated views must retain fine detail, follow the requested camera poses, and depict consistent geometry and appearance across overlapping regions~\cite{worldstereo2026}.
Otherwise, conflicting observations can produce blurred textures or distorted structures when fitted to a single 3D scene.

Learning to produce these observations also depends on suitable panoramic training data.
Existing collections vary in spatial resolution and visual quality and provide uneven coverage across scene categories~\cite{dvd3602024,panowan2025}.
Although recent datasets provide camera trajectories and geometric annotations~\cite{matrix3d2025,omniroam2026,panoworld_real2026}, the availability of data that jointly support high-resolution, high-quality generation across diverse environments remains limited.
This matters particularly for panoramas, whose pixels are distributed over the full sphere: limited resolution leaves little detail in individual viewing directions.
Degraded imagery and restricted scene coverage further limit the appearance and structural diversity available for learning.

To address these data and generation challenges, we present \textbf{PanoTrek}, a framework that reconstructs an explicit 3D Gaussian scene from an input panorama using generated panoramic videos as multi-view observations (Fig.~1).
We first construct \textbf{PanoTraj}, a dataset of over 20,000 rendered panoramic video clips from more than 1,000 indoor and outdoor scenes.
It provides sequences at resolutions up to $2880 \times 1440$, paired with calibrated camera trajectories and aligned geometric conditions.
These sequences supervise panoramic video adaptation, while the matched
trajectories and geometric conditions support camera-control learning.

Using PanoTraj, we train the video generator in two stages:
\textbf{Panoramic Video Adaptation (PVA)} and
\textbf{Geometry-Guided Camera Control (GCC)}.
PVA adapts a pretrained video prior to panoramic video generation
through low-rank adaptation (LoRA) and boundary-aware supervision.
GCC then learns camera control while keeping the adapted prior fixed.
A dedicated control branch receives coarse renderings of the input
panorama's geometry at the requested viewpoints, together with
confidence maps that indicate their reliability.
These conditions guide the model to preserve supported scene content
while refining distorted projections and completing newly exposed regions.

As generation extends across video chunks, overlapping regions must remain compatible with the same underlying scene.
We introduce \textbf{3DCache}, a mesh-based geometric memory that retains previously generated scene content and reprojects it to guide subsequent synthesis.
This reuse constrains later views with established geometry and appearance, reducing conflicts between separately generated chunks.
The resulting panoramas are converted into perspective observations and used with their prescribed camera poses to optimize a 3D Gaussian Splatting representation~\cite{kerbl2023gaussians}.
Video generation thus supplies the observations for reconstruction, while the resulting 3D scene supports novel-view rendering.

Experiments demonstrate state-of-the-art panoramic video fidelity
and camera controllability on PanoTraj, with reconstructed scenes
exhibiting substantially higher rendering quality than those of
competing methods.
Evaluations with two pretrained backbones further show compatibility with existing few-step generators.
Our main contributions are as follows:
\begin{itemize}
    \item To support high-resolution panoramic generation for 3D reconstruction, we construct PanoTraj, providing indoor and outdoor video sequences with calibrated trajectories and aligned geometric conditions at resolutions up to $2880 \times 1440$.
    \item To synthesize detailed observations aligned with prescribed viewpoints, we develop a two-stage framework that adapts pretrained video priors to panoramic video generation and learns camera control through geometric conditioning.
    \item To maintain consistent reconstruction observations across video chunks and trajectories, we introduce 3DCache, a mesh-based memory that reuses previously generated scene geometry and appearance to guide subsequent synthesis.
\end{itemize}

\section{Related Work}
\label{sec:related_work}

\paragraph{Video generation.}
Large-scale video diffusion models provide strong visual and temporal priors for text- and image-conditioned synthesis~\cite{svd2023,cogvideox2024,hunyuanvideo2024,wan2025}.
Camera-controlled models introduce pose or ray conditioning~\cite{motionctrl2024,cameractrl2025,vd3d2025,ac3d2025}, while epipolar attention strengthens correspondence across viewpoints~\cite{camco2024,cami2v2024}.
Related video re-rendering methods redirect the trajectory of existing footage while preserving scene appearance and dynamics~\cite{recammaster2025,trajectorycrafter2025}.
Rendering reconstructed geometry into target viewpoints provides spatially aligned guidance for generation~\cite{uni3c2025,nvssolver2025,viewcrafter2025}.
Continued scene exploration extends this setting beyond individual clips~\cite{cameractrlii2025}, motivating complementary mechanisms for retaining and reusing generated content.
Explicit 3D memories accumulate scene geometry and appearance and reproject them to guide subsequent viewpoints~\cite{gen3c2025,worldstereo2026}.
History-retrieval methods supply earlier frames or latents selected for their relevance to the target view as additional generation context~\cite{contextmemory2025,solarwmmemory2026}.
Implicit approaches retain history within temporal context or compressed latent caches~\cite{lingbotworld2026,relic2025}, or encode retrieved observations into learned 3D-aware representations queried by target camera poses~\cite{worldcrafter2026}.
PanoTrek combines explicit geometric guidance and shared scene memory with panoramic adaptation to constrain generation across video chunks.

\paragraph{Panoramic image and video generation.}
Panoramic image generation establishes omnidirectional scene context from text prompts~\cite{panogen2023,stitchdiffusion2024,panfusion2024,unipano2025} or expands perspective observations into complete panoramas~\cite{diffusion3602023,panodiffusion2024}.
Recent diffusion transformers further improve text-conditioned synthesis and image-conditioned outpainting~\cite{dit3602026,hyworld2026}.
Geometric correspondences and shared spatial context support consistent synthesis across viewing directions and camera positions~\cite{mvdiffusion2023,diffpano2024,panoworld_house2026}.
Extending panoramic generation to video requires preserving both spherical continuity and temporal coherence.
Existing approaches adapt pretrained video priors to panoramic video generation~\cite{dvd3602024,panolora2025,panowan2025} or expand perspective videos through coordinated multi-view generation and outpainting~\cite{argus2025,imagine3602025,videopanda2025}.
Spatially scalable generation supports high-resolution panoramic dynamics~\cite{fourk4dgen2024,dynamicscaler2025}, while depth and tracked-point supervision strengthen geometric consistency over time~\cite{panoworld_geometry2026}.
For scene exploration, action conditioning and explicit camera guidance enable panoramic synthesis along controlled trajectories~\cite{genex2025,campvg2025,matrix3d2025,panoworldx2025}.
Temporal refinement and scene memory further support extended exploration~\cite{omniroam2026,panoworld_real2026,evoworld2025}.
PanoTrek combines panoramic adaptation with geometry-guided camera control to produce high-fidelity panoramic videos at native resolutions up to $2880 \times 1440$.

\paragraph{Generative 3D scene reconstruction.}
Generative reconstruction combines visual priors with 3D scene representations to infer content beyond the available observations.
Early approaches build scenes through image synthesis, depth estimation, and geometric integration~\cite{luciddreamer2025,text2room2023,wonderjourney2024,wonderworld2025}, while view-conditioned diffusion priors can also supervise scene optimization through score distillation~\cite{zeronvs2024}.
Recent methods connect generation and reconstruction in several ways.
One route couples generation with 3D prediction by decoding Gaussian scenes from video latents~\cite{lyra2026,wonderland2025,videorfsplat2025} or jointly modeling aligned appearance and geometric latents~\cite{gen3r2026}.
Within this route, other approaches couple Gaussian prediction with diffusion and differentiable rendering~\cite{diffusiongs2025,flashworld2026,pixworld2026}.
A second route synthesizes additional views from single or sparse observations and reconstructs them into a scene, using multi-view image priors~\cite{cat3d2024} or video priors~\cite{reconx2026,dimensionx2025}.
Geometric memory and history retrieval extend this approach to continued scene expansion~\cite{lyra2_2026,hyworld2026}.
Panoramic pipelines further broaden spatial coverage: geometric lifting, RGB-D completion, and layered representations construct scenes from a panorama~\cite{dreamscene3602024,pano2room2024,layerpano3d2024,scene4u2025}, while panoramic video provides observations from multiple camera positions for subsequent reconstruction~\cite{worldprompter2025,matrix3d2025,omniroam2026,geniesim_panoworld2026}.
PanoTrek follows this panoramic video route, using high-resolution observations and cross-chunk geometric memory to improve the consistency of the views used for 3D Gaussian reconstruction.
\begin{figure*}
    \centering
    \includegraphics[width=1\linewidth]{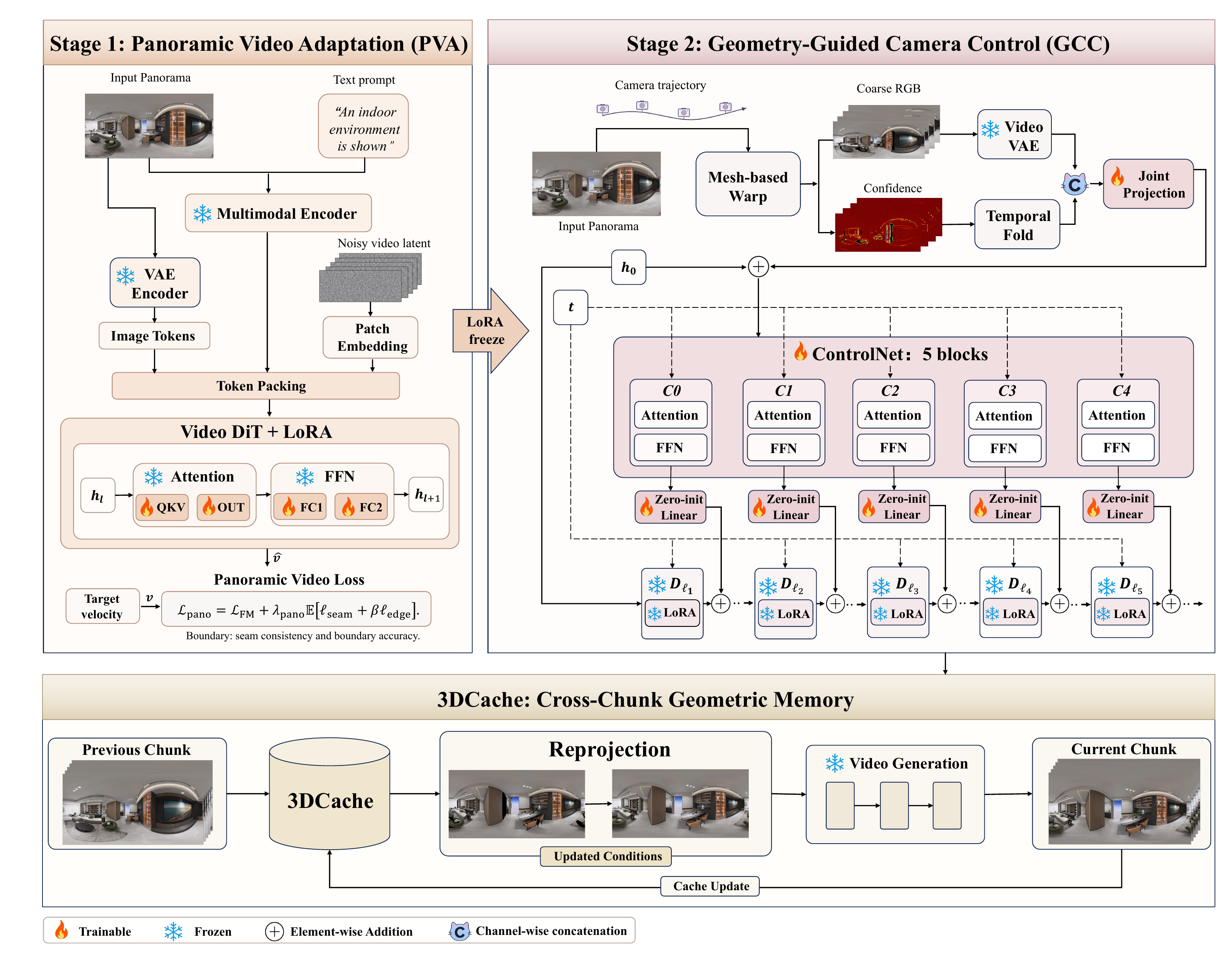}
    \caption{\textbf{Overview of PanoTrek.}
    Panoramic Video Adaptation (PVA) adapts a pretrained video prior to panoramic video generation through LoRA and boundary-aware supervision.
    Geometry-Guided Camera Control (GCC) learns a control branch conditioned
    on coarse RGB renderings and projection confidence, while keeping the
    adapted backbone and LoRA weights frozen.
    At inference, 3DCache reuses historical scene content as geometric
    conditioning and denoising guidance to improve consistency across
    video chunks and camera trajectories.}
    \label{fig:gaussworld_framework}

\end{figure*}
\section{Method}
\label{sec:method}

Given an input panorama $I_0$ and camera-to-world poses
$\mathcal{T}=\{T_i\}_{i=0}^{N-1}$, PanoTrek generates panoramic videos
as multi-view observations for reconstructing an explorable 3D scene
(Fig.~\ref{fig:gaussworld_framework}).
Our training consists of PVA followed by GCC, adapting the video prior to panoramic video generation before learning camera control from coarse
geometric renderings and their confidence maps.
At inference, 3DCache reuses textured meshes from earlier generations
to maintain geometric consistency across video chunks.
The generated panoramas are converted into perspective observations
and used with their prescribed camera poses to reconstruct a 3D Gaussian
Splatting representation for novel-view rendering.

\subsection{Panoramic Video Adaptation}
\label{sec:gw_adaptation}

Pretrained video models are primarily trained on perspective videos, whereas equirectangular projection (ERP) introduces
latitude-dependent distortion and horizontal periodicity.
We therefore adapt the model to panoramic videos using
low-rank adaptation (LoRA)~\cite{hu2022lora}, while keeping
the pretrained weights frozen.

For a noisy video latent $z_\tau$ at flow time $\tau$,
let $\hat{v}_\tau$ and $v_\tau$ denote the predicted and target
flow velocities.
The flow-matching objective minimizes their mean squared error:
\begin{equation}
\mathcal{L}_{\mathrm{FM}}
=\mathbb{E}\!\left[
\operatorname{MSE}(\hat{v}_\tau,v_\tau)
\right],
\label{eq:gw_fm_loss}
\end{equation}
where the expectation is over training examples, noise, and flow time.
To account for the horizontal periodicity of ERP, we additionally
supervise prediction consistency across the two boundaries
and prediction accuracy in their vicinity:
\begin{equation}
\begin{aligned}
\ell_{\mathrm{seam}}
    &= \operatorname{SmoothL1}\!\left(
        \hat{v}_\tau^L-\hat{v}_\tau^R,
        v_\tau^L-v_\tau^R
    \right),\\
\ell_{\mathrm{edge}}
    &= \operatorname{MSE}\!\left(
        \hat{v}_\tau|_{\mathcal{B}},
        v_\tau|_{\mathcal{B}}
    \right),\\
\mathcal{L}_{\mathrm{pano}}
    &= \mathcal{L}_{\mathrm{FM}}
    +\lambda_{\mathrm{pano}}\mathbb{E}\!\left[
        \ell_{\mathrm{seam}}+\beta\ell_{\mathrm{edge}}
    \right].
\end{aligned}
\label{eq:gw_pano_loss}
\end{equation}
Here, $L$ and $R$ denote the leftmost and rightmost latent
columns, and $\mathcal{B}$ is a narrow band along these boundaries.
The seam term matches the predicted cross-boundary variation
to the target, while the edge term strengthens local prediction accuracy.

\subsection{Geometry-Guided Camera Control}
\label{sec:gw_camera_control}

After PVA, GCC learns to follow specified camera trajectories using
coarse RGB renderings and their confidence maps as conditions.
Coarse RGB depicts the observed scene from the requested viewpoints,
providing spatial guidance for camera motion.
Confidence indicates where this guidance is reliable and where
projection-induced texture deformation calls for refinement.

\paragraph{Coarse RGB rendering.}
We lift $I_0$ into a textured triangle mesh $\mathcal{M}_0$
using radial depth derived from MoGe-3~\cite{moge32026} estimates.
Valid pixels are back-projected along spherical rays and connected
to neighboring samples, excluding connections across depth discontinuities.
The mesh is expressed in the input camera frame and textured with $I_0$.
For each target pose $T_i$, the coarse RGB rendering is
\begin{equation}
\tilde{I}_i
=\mathcal{R}\!\left(\mathcal{M}_0,T_i^{-1}T_0\right),
\label{eq:gw_mesh_warp}
\end{equation}
where $T_0$ is the camera-to-world pose of the input panorama
and $\mathcal{R}$ transforms and rasterizes the mesh in ERP.
The renderings form $\tilde{V}=\{\tilde{I}_i\}_{i=0}^{N-1}$;
$\Omega_i^{\mathrm{obs}}$ denotes their valid visible pixels.

\paragraph{Projection confidence.}
Reprojection can stretch source textures even where geometry is visible.
Let $p$ denote a pixel location in a target panorama and
$\Omega^{\mathrm{obs}}$ the set of pixels supported by valid visible
mesh projections; we omit the frame index for clarity.
For $p\in\Omega^{\mathrm{obs}}$, let $w(p)$ denote the corresponding
source-panorama location.
We measure local deformation through $J(p)=\partial w(p)/\partial p$
in spherical tangent coordinates, accounting for ERP distortion.
Its singular values $\sigma_1(p)\geq\sigma_2(p)>0$ give the maximum
magnification $m(p)=1/\sigma_2(p)$ and anisotropy
$a(p)=\sigma_1(p)/\sigma_2(p)$.
The penalties beyond tolerances $\tau_{\mathrm{mag}}$ and
$\tau_{\mathrm{ani}}$ are
\begin{equation}
\begin{aligned}
r_{\mathrm{mag}}(p)
    &= \left[\log\!\left(m(p)/\tau_{\mathrm{mag}}\right)\right]_+,\\
r_{\mathrm{ani}}(p)
    &= \left[\log\!\left(a(p)/\tau_{\mathrm{ani}}\right)\right]_+,
\end{aligned}
\label{eq:gw_deformation}
\end{equation}
where $[x]_+=\max(x,0)$.
Combining these with nonnegative depth-discontinuity and
occlusion-boundary penalties $r_{\mathrm{edge}}(p)$ and
$r_{\mathrm{occ}}(p)$ yields
\begin{equation}
\begin{aligned}
R(p) &= r_{\mathrm{mag}}^2(p)
       +\lambda_{\mathrm{ani}}r_{\mathrm{ani}}^2(p)\\
     &\quad +\lambda_{\mathrm{edge}}r_{\mathrm{edge}}(p)
       +\lambda_{\mathrm{occ}}r_{\mathrm{occ}}(p),\\
C(p) &=
\begin{cases}
\exp\!\left(-R(p)\right), & p\in\Omega^{\mathrm{obs}},\\
0, & \text{otherwise}.
\end{cases}
\end{aligned}
\label{eq:gw_confidence}
\end{equation}
Here, $\lambda_{\mathrm{ani}}$, $\lambda_{\mathrm{edge}}$, and
$\lambda_{\mathrm{occ}}$ are fixed nonnegative coefficients that
balance the penalties relative to magnification.
The confidence map assigns zero to unsupported pixels and lower values
to projections with stronger deformation or boundary artifacts.

\paragraph{Conditioning and training.}
The frozen video VAE $\mathcal{E}$ encodes the coarse RGB sequence,
and $\mathcal{A}$ aligns confidence to the same latent grid.
We concatenate the two conditions along channels ($\|$),
patchify them, and apply a learned projection $\mathcal{P}$:
\begin{equation}
q=\mathcal{P}\!\left(
\operatorname{Patch}\!\left[
\mathcal{E}(\tilde{V})\,\|\,\mathcal{A}(C)
\right]\right).
\label{eq:gw_condition_tokens}
\end{equation}
A ControlNet-style branch~\cite{zhang2023controlnet}, initialized
from selected backbone blocks, processes these condition tokens
alongside the noisy video tokens and injects residuals through
zero-initialized projections.
We optimize only $\mathcal{P}$ and the control branch using
the flow-matching objective in Eq.~\eqref{eq:gw_fm_loss},
with the panorama-adapted backbone and LoRA weights frozen.
The model also retains its input-panorama and text conditioning.

\subsection{3DCache for Scene Expansion}
\label{sec:gw_cache}

Extending exploration across video chunks requires later generations
to respect previously established scene geometry and appearance.
We introduce \textbf{3DCache}, a mesh-based geometric memory
that provides this shared reference.

\paragraph{Mesh-based memory.}
After each chunk, we align estimated depths with available reference
depths and construct textured triangle meshes from selected generated
panoramas.
Each mesh retains its source camera pose, placing historical surfaces
in a shared coordinate system, with connections excluded across
depth discontinuities.
For subsequent chunks, we render selected meshes at the target poses
to supplement unobserved regions of the coarse RGB conditions.
Confidence is updated according to valid projected coverage,
so partially observed regions receive proportionally weaker support.

\paragraph{Memory-guided generation.}
To preserve the reprojected appearance during synthesis, we also
constrain the denoising process.
Using the updated RGB and confidence conditions, we first generate
a provisional video that supplies context for the constraint.
In newly cache-supported regions whose original confidence was zero,
we composite projected textures into this video and encode the result
with the frozen VAE as a target latent $z^{\mathrm{tar}}$.
A second sampling pass guides each predicted clean latent:
\begin{equation}
\hat{z}^{+}
=\hat{z}+\alpha M\odot\left(z^{\mathrm{tar}}-\hat{z}\right),
\label{eq:gw_cache_guidance}
\end{equation}
where $M$ carries the coverage-aware confidence to the latent grid,
is restricted to these newly supported regions, and is zero for
the first frame; $\alpha$ controls guidance strength.
The corrected clean latent determines the velocity used in each
sampling update.
The resulting panoramas are converted into perspective observations
for 3DGS reconstruction using their prescribed camera poses.

\subsection{PanoTraj Dataset}
\label{sec:gw_data}

We construct \textbf{PanoTraj}, a dataset of over 20,000
panoramic video clips from more than 1,000 indoor and outdoor scenes.
The source environments comprise publicly available
InteriorGS~\cite{interiorgs2025,miao2026towards} and
Replica~\cite{replica2019} scenes, together with our
in-house BlockGS scenes.
InteriorGS provides 3DGS environments and Replica provides
textured reconstructions; for BlockGS, we reconstruct 3DGS scenes
from captured videos.
We sample camera trajectories within these environments
and render panoramic RGB and depth sequences with known camera parameters.
Each clip is paired with coarse RGB and confidence conditions
constructed from its initial panorama and target trajectory.
Panoramic sequences supervise PVA, while their calibrated trajectories
and aligned geometric conditions support GCC.
Scenes are partitioned into disjoint training and test sets.
In addition to $1440\times720$ videos, PanoTraj includes
$2880\times1440$ sequences for high-resolution generation with
the same training procedure.
\begin{figure*}[t]
    \centering
    \includegraphics[width=0.88\textwidth]{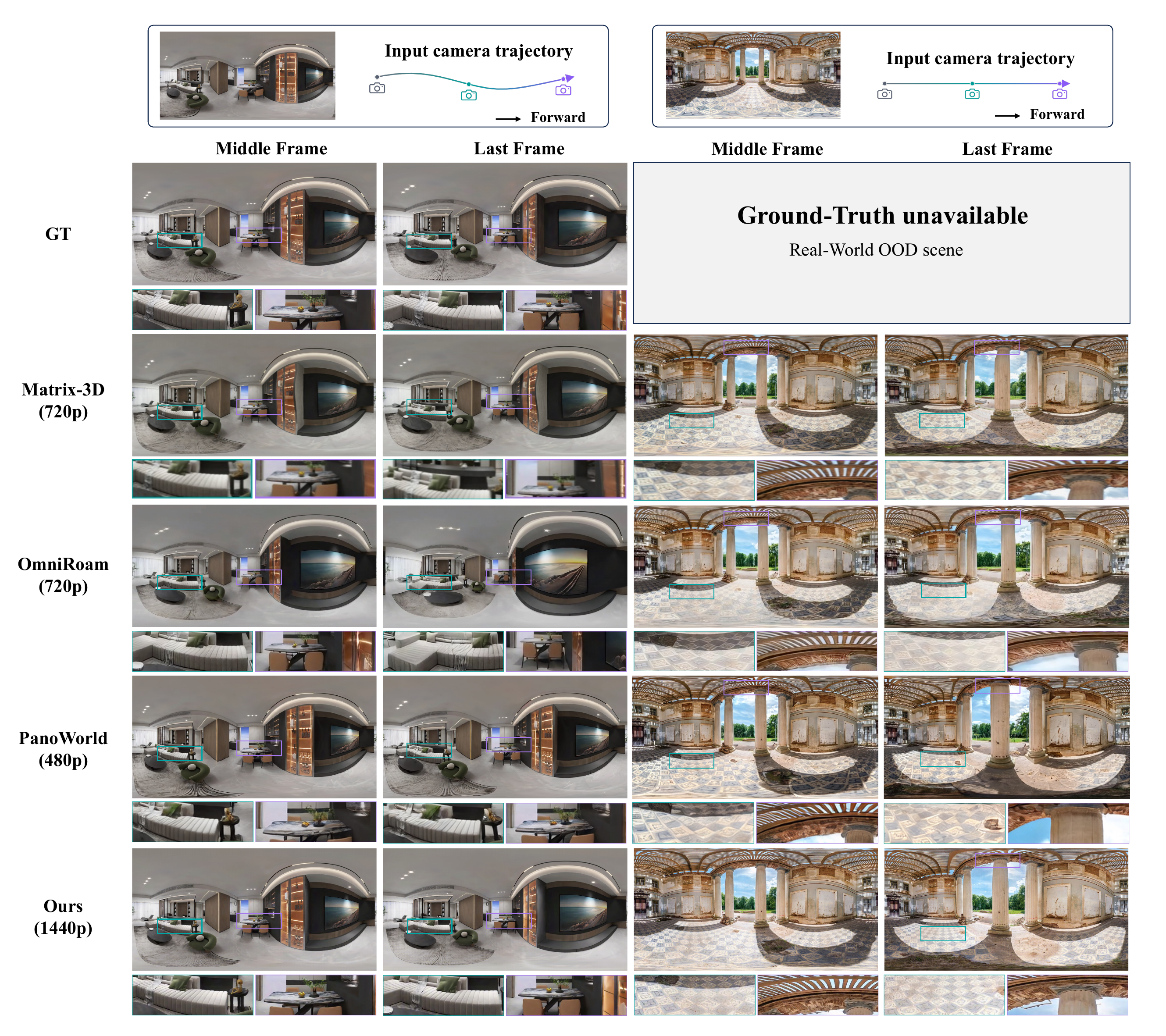}
    \caption{\textbf{Camera-controlled panoramic video generation.}
    Intermediate and final views are shown with enlarged details.
    PanoTrek preserves finer visual details and more coherent
    scene structures during camera motion, including on a real-world
    out-of-distribution input.
    Ground-truth target views are available for the left scene only.
    Labels indicate each method's native output resolution.}
    \label{fig:gw_qualitative}
\end{figure*}

\section{Experiments}
\label{sec:experiments}

\subsection{Experimental Setup}
\label{sec:exp_setup}

\paragraph{Backbones and training.}
We instantiate PanoTrek on Wan 2.1 I2V 14B~\cite{wan2025}
and MiniMax-H3 FL2VA 33B~\cite{h32026}, generating panoramic
videos at $1440\times720$ and $2880\times1440$ with 21 and
22 frames, respectively. PVA uses LoRA with rank 128.
GCC initializes its control branch by copying five DiT blocks
from the pretrained backbone and zero-initializes the residual
output projections. The panorama-adapted backbone and PVA
LoRA remain frozen during GCC. Both stages are trained for
2,500 iterations on 16 NVIDIA A800 GPUs.

\paragraph{Inference settings.}
Inference is performed on NVIDIA A800 GPUs, with standard
sampling using 50 denoising steps per pass.
To evaluate compatibility with pretrained few-step generators,
we additionally use the rank-64 CFG- and step-distilled
LightX2V adapter~\cite{lightx2v2025} for Wan and the
full-critic DMAD checkpoint~\cite{dmad2026} with an INT8
H3 backbone.

\begin{table*}[t]
\centering
\caption{Panoramic video generation on 200 held-out clips at native output
resolutions. Best and second-best quality and camera-control scores are
bold and underlined. $\dagger$ Imagine360 uses ground-truth perspective
video and does not support explicit camera control.}
\label{tab:gw_video_generation}
\begingroup
\rmfamily\fontsize{7.3}{8.8}\selectfont
\setlength{\tabcolsep}{2.2pt}
\begin{tabular*}{\textwidth}{@{\extracolsep{\fill}}lc*{11}{r}@{}}
\toprule
& & \multicolumn{3}{c}{Fidelity} & \multicolumn{5}{c}{Visual \& Video Quality}
& \multicolumn{2}{c}{Camera Control} & \\
\cmidrule(lr){3-5}\cmidrule(lr){6-10}\cmidrule(lr){11-12}
Method & Res. & PSNR (dB)$\uparrow$ & SSIM$\uparrow$ & LPIPS$\downarrow$
& FID$\downarrow$ & FVD$\downarrow$ & NIQE$\downarrow$
& $\mathrm{QA}_q\uparrow$ & $\mathrm{QA}_a\uparrow$
& $R_{\mathrm{err}}$ (rad)$\downarrow$ & $T_{\mathrm{err}}\downarrow$ & Time (s)$\downarrow$ \\
\midrule
Imagine360$^{\dagger}$ & 480p & 14.701 & 0.622 & 0.544 & 95.715 & 1042.191 & \textbf{4.318} & 3.6299 & 3.3494 & -- & -- & 217 \\
Matrix-3D & 480p & 19.375 & 0.745 & 0.344 & 24.065 & 371.075 & 6.146 & 3.5376 & 3.3097 & 0.03438 & 0.10951 & 1058 \\
OmniRoam (Preview) & 480p & 17.195 & 0.699 & 0.383 & 25.198 & 490.882 & \underline{4.710} & 4.1081 & 3.8776 & 0.05809 & 0.13865 & 651 \\
PanoWorld & 480p & 20.320 & 0.750 & 0.230 & 18.333 & 136.961 & 5.683 & 4.1963 & 3.9065 & 0.02021 & 0.25891 & 86 \\
\midrule
Matrix-3D & 720p & 19.278 & 0.774 & 0.351 & 23.654 & 402.204 & 6.199 & 4.0008 & 3.7512 & 0.04372 & 0.11616 & 3948 \\
OmniRoam (Refine) & 720p & 16.844 & 0.740 & 0.407 & 29.629 & 537.424 & 5.819 & 4.2097 & 3.9604 & 0.07519 & 0.16139 & 2850 \\
Ours (Wan) & 720p & \underline{26.877} & \textbf{0.888} & \underline{0.091} & \underline{7.622} & \textbf{54.115} & 5.234 & 4.2894 & 4.0336 & \underline{0.01468} & \underline{0.10034} & 718 \\
Ours (Wan + LightX2V) & 720p & \textbf{26.983} & \underline{0.887} & \textbf{0.087} & \textbf{7.532} & \underline{56.457} & 5.259 & \textbf{4.3212} & \textbf{4.0655} & \textbf{0.01284} & \textbf{0.09516} & 29 \\
\midrule
Ours (H3) & 1440p & 24.229 & 0.814 & 0.210 & 26.542 & 93.900 & 6.112 & 4.0051 & 3.7740 & 0.02943 & 0.15988 & 934 \\
Ours (H3 + DMAD) & 1440p & 23.389 & 0.863 & 0.156 & 10.025 & 72.252 & 5.762 & \underline{4.3162} & \underline{4.0398} & 0.01619 & 0.14072 & 79 \\
\bottomrule
\end{tabular*}
\endgroup
\end{table*}

\paragraph{Evaluation dataset and metrics.}
We evaluate on 200 test clips from PanoTraj, using
a scene-disjoint training and test split.
All methods are evaluated at their native output resolutions.
We measure reference fidelity using PSNR, SSIM~\cite{ssim2004},
and LPIPS~\cite{lpips2018}, and distributional quality using
FID~\cite{fid2017} and FVD~\cite{fvd2018}.
NIQE~\cite{niqe2013} and Q-Align~\cite{qalign2024} assess
perceptual quality, with $\mathrm{QA}_q$ and $\mathrm{QA}_a$
denoting quality and aesthetic scores, respectively.
For camera control, we report rotation error $R_{\mathrm{err}}$
in radians and normalized translation error $T_{\mathrm{err}}$,
using poses estimated by VGGT~\cite{vggt2025} from four
perspective views of each panorama.

\paragraph{Baselines.}
We compare against Matrix-3D~\cite{matrix3d2025},
OmniRoam~\cite{omniroam2026}, and
PanoWorld~\cite{panoworld_real2026}.
We also include Imagine360~\cite{imagine3602025} for
perspective-video outpainting.
It receives ground-truth perspective videos rather than
a panorama and a camera trajectory, and is excluded
from the camera-control evaluation.

\subsection{Evaluation on Panoramic Video Generation}
\label{sec:gw_video_eval}

\paragraph{Visual fidelity and quality.}
Table~\ref{tab:gw_video_generation} compares panoramic video
generation at native output resolutions.
At 720p, both Wan-based variants outperform Matrix-3D and
OmniRoam in PSNR, SSIM, LPIPS, FID, and FVD.
Compared with Matrix-3D at the same resolution, our
standard Wan variant increases PSNR from 19.278 to
26.877\,dB and reduces LPIPS from 0.351 to 0.091,
showing improved reference fidelity in both pixel and
perceptual comparisons.
The accelerated Wan variant also obtains the highest
Q-Align quality and aesthetic scores in the benchmark.

\paragraph{Camera controllability.}
PanoTrek achieves state-of-the-art camera controllability
in Table~\ref{tab:gw_video_generation}, with both Wan-based
variants outperforming all competing methods in rotation
and translation accuracy.
The LightX2V variant obtains the lowest rotation and
translation errors of 0.01284 and 0.09516, respectively,
while the standard Wan variant ranks second on both
metrics at 0.01468 and 0.10034.
At 1440p, DMAD reduces the H3 variant's rotation error
from 0.02943 to 0.01619 and translation error from
0.15988 to 0.14072, demonstrating that camera control
is retained when using an accelerated backbone.

\paragraph{Compatibility with accelerated inference.}
Pretrained acceleration models substantially reduce inference
time for both implementations.
LightX2V reduces Wan runtime from 718 to 29 seconds per clip,
while DMAD reduces H3 runtime from 934 to 79 seconds,
corresponding to speedups of $24.8\times$ and $11.8\times$,
respectively.
Wan retains comparable fidelity and video quality under
acceleration.
For H3, DMAD improves SSIM, LPIPS, FID, and FVD, although
PSNR decreases from 24.229 to 23.389\,dB.
These results demonstrate that PanoTrek can accommodate
existing few-step video generators across both backbones.

\paragraph{Qualitative comparison and generalization.}
Fig.~\ref{fig:gw_qualitative} shows that PanoTrek preserves
finer visual details and more coherent scene structures
as the viewpoint changes.
The real-world out-of-distribution comparison further shows
clearer textures and more consistent scene appearance,
providing qualitative evidence of generalization beyond
the rendered training scenes.

\begin{figure*}
    \centering
    \includegraphics[width=1.0\linewidth]{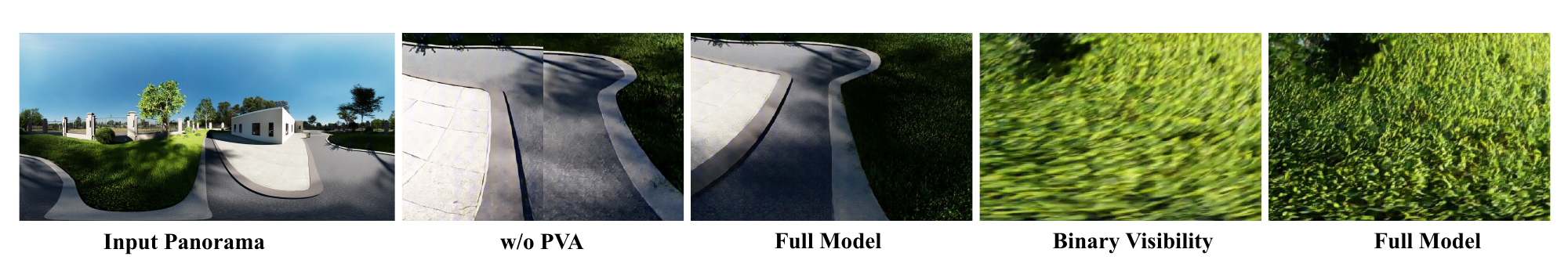}
    \caption{\textbf{Effects of PVA and confidence conditioning in GCC.}
    After the input panorama, the middle pair compares generation without
    and with PVA, showing improved continuity across the panoramic boundary.
    The right pair compares binary visibility with continuous projection
    confidence in GCC, which better recovers high-frequency textures
    affected by severe projection stretching.}
    \label{fig:gw_ablation}
\end{figure*}

\begin{table}[t]
\centering
\caption{Rendering quality on three randomly selected test scenes.
Res. is the generation resolution; Gaussian counts are scene averages
in millions. Best quality scores are in bold.}
\label{tab:gw_reconstruction}
\begingroup
\rmfamily\fontsize{7.5}{9.25}\selectfont
\setlength{\tabcolsep}{2pt}
\begin{tabular*}{\columnwidth}{@{\extracolsep{\fill}}lcrrrr@{}}
\toprule
Method & Res. & PSNR (dB)$\uparrow$ & SSIM$\uparrow$ & LPIPS$\downarrow$ & Gaussians (M) \\
\midrule
Matrix-3D & 720p & 24.561 & \textbf{0.875} & 0.247 & 0.4036 \\
OmniRoam & 720p & 23.654 & 0.849 & 0.237 & 0.7854 \\
PanoWorld & 480p & 19.854 & 0.771 & 0.376 & 0.9354 \\
\midrule
Ours (Wan) & 720p & 25.068 & 0.860 & 0.198 & 0.8649 \\
Ours (H3) & 1440p & \textbf{25.078} & 0.863 & \textbf{0.177} & 1.1663 \\
\bottomrule
\end{tabular*}
\endgroup
\end{table}

\subsection{Evaluation on 3D Scene Reconstruction}
\label{sec:gw_reconstruction}

We randomly select three test scenes and reconstruct each method's
outputs using the same iterative 3DGS pipeline for 30,000 iterations,
without capping the Gaussian count.
For each scene, we use 11 uniformly sampled frames from the generated
chunk for fitting and 10 held-out frames for evaluation.
Each panorama is converted into 24 perspective views, yielding
240 test images per scene.
Our 720p and 1440p variants use the standard Wan and H3 models, respectively.

Table~\ref{tab:gw_reconstruction} shows that PanoTrek at 720p achieves
25.068\,dB PSNR and 0.198 LPIPS, outperforming all baselines on both metrics.
The 1440p H3 variant further reduces LPIPS to 0.177, while Matrix-3D
retains the highest SSIM.
These results support the effectiveness of our generated observations
for downstream reconstruction.
Additional visualizations of the reconstructed scenes are provided
in the supplementary material.

\subsection{Ablation Studies}
\label{sec:gw_ablation}

\begin{table}[t]
\centering
\caption{Comparison and ablation on challenging held-out scenes.
Our H3 variants use 1440p; ablations are trained separately
without PVA or with binary visibility in GCC.
Best scores are in bold.}
\label{tab:gw_ablation}
\begingroup
\rmfamily\fontsize{7.5}{9.25}\selectfont
\setlength{\tabcolsep}{2pt}
\begin{tabular*}{\columnwidth}{@{\extracolsep{\fill}}lrrrr@{}}
\toprule
Method / Variant & PSNR (dB)$\uparrow$ & SSIM$\uparrow$ & LPIPS$\downarrow$ & FID$\downarrow$ \\
\midrule
Matrix-3D (720p) & 12.994 & 0.447 & 0.517 & 81.226 \\
OmniRoam (720p) & 13.773 & 0.442 & 0.432 & 86.687 \\
\midrule
Ours (binary visibility) & 15.334 & \textbf{0.473} & 0.342 & 58.912 \\
Ours (w/o PVA) & 13.503 & 0.421 & 0.390 & 64.448 \\
Ours (full) & \textbf{15.529} & 0.468 & \textbf{0.308} & \textbf{53.550} \\
\bottomrule
\end{tabular*}
\endgroup
\end{table}

We evaluate the effect of PVA and the role of confidence conditioning
in GCC using the H3-based model at 1440p.
Both ablated variants are trained separately and evaluated on challenging
held-out clips with large camera displacements and rich high-frequency
textures.
We also assess the effect of 3DCache on cross-chunk consistency.

\paragraph{Effect of PVA.}
To assess PVA, we compare the full model with a separately trained
variant that omits panoramic adaptation while retaining confidence
conditioning.
Removing PVA degrades all four metrics in Table~\ref{tab:gw_ablation}:
PSNR drops from 15.529 to 13.503 dB, and LPIPS increases from 0.308 to 0.390.
Fig.~\ref{fig:gw_ablation} further shows discontinuities
at the panoramic boundary without adaptation, whereas
the full model maintains better boundary continuity.
These results support panoramic adaptation for improving
both visual fidelity and panoramic boundary consistency.

\paragraph{Confidence conditioning in GCC.}
To evaluate confidence conditioning in GCC, we replace continuous
projection confidence with binary visibility while retaining panoramic
adaptation and camera-control training.
Large camera displacements can severely stretch high-frequency
textures in these challenging scenes.
Binary visibility still treats the distorted textures as
valid conditioning evidence.
As shown in Fig.~\ref{fig:gw_ablation}, the binary-visibility variant
retains the stretched patterns, while continuous confidence
enables the model to regenerate finer textures in unreliable
regions.
Table~\ref{tab:gw_ablation} shows corresponding improvements
in PSNR, LPIPS, and FID over binary conditioning, with LPIPS
decreasing from 0.342 to 0.308 and FID from 58.912 to 53.550.
SSIM decreases slightly, from 0.473 to 0.468.

\begin{figure}[t]
    \centering
    \includegraphics[width=\columnwidth]{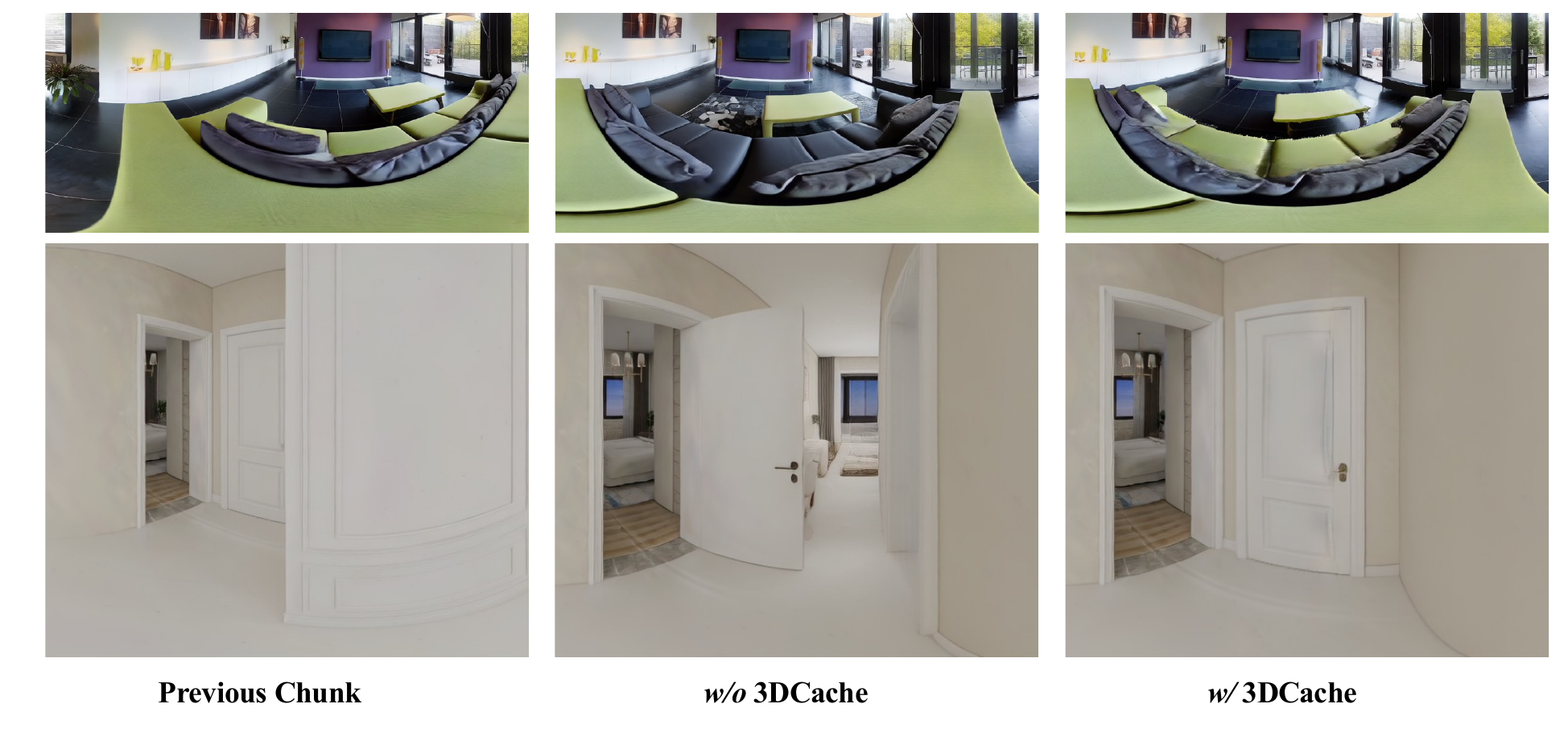}
    \caption{\textbf{Effect of 3DCache on cross-chunk consistency.}
    Previous observations (left) are compared with subsequent
    generation without (middle) and with (right) 3DCache.
}
    \label{fig:gw_3dcache_ablation}
\end{figure}

\paragraph{Cross-chunk consistency.}
Fig.~\ref{fig:gw_3dcache_ablation} compares successive chunks
in overlapping regions.
Without 3DCache, later generations introduce appearance and
structural inconsistencies with earlier content.
3DCache better preserves scene structure and appearance
across chunks for coherent expansion.
Reconstruction results for the 3DCache ablation are provided
in the supplementary material.


\section{Conclusion}
\label{sec:conclusion}

We presented PanoTrek, a framework that generates camera-controlled
panoramic videos as observations for reconstructing explorable 3D scenes.
Our two-stage training adapts pretrained video priors to panoramic video generation and learns viewpoint control from coarse
renderings and their confidence.
3DCache preserves established scene content across video
chunks, supporting consistent scene expansion.
We also introduced PanoTraj, a panoramic video dataset
with calibrated camera trajectories.
Experiments demonstrate state-of-the-art panoramic video
fidelity and camera controllability on PanoTraj, alongside detailed
3D reconstruction from high-resolution observations.
Evaluations on two backbones further show compatibility
with pretrained few-step generators.
These findings establish controllable panoramic generation
with persistent geometric memory as an effective approach
to constructing coherent, explorable 3D scenes.

\section*{LLM Usage Statement}
We used OpenAI's ChatGPT to assist with manuscript writing and editing, including improving clarity, refining academic language, and organizing the presentation of related work, methodology, and experimental findings. All scientific contributions, experimental results, and final conclusions were developed and verified by the authors.

{\small\bibliographystyle{ieeenat_fullname}\bibliography{main}}
\clearpage
\appendix
\setcounter{figure}{0}
\setcounter{table}{0}
\setcounter{equation}{0}
\renewcommand{\thefigure}{S\arabic{figure}}
\renewcommand{\thetable}{S\arabic{table}}
\renewcommand{\theequation}{S\arabic{equation}}
\renewcommand{\theHfigure}{supp.\arabic{figure}}
\renewcommand{\theHtable}{supp.\arabic{table}}
\renewcommand{\theHequation}{supp.\arabic{equation}}
\maketitlesupplementary
\noindent This supplementary material expands the method and implementation details of PanoTrek, describes the dataset construction pipeline, and analyzes additional video generation and 3D reconstruction results.
\begin{itemize}
\setlength{\itemsep}{1pt}
\setlength{\parskip}{0pt}
\item \textbf{Section~\ref{supp:method_details}: Method Details.} Input-to-mesh construction, coarse video conditioning, and geometric memory reuse.
\item \textbf{Section~\ref{supp:training}: Implementation Details.} Model configuration, training, inference, and evaluation protocols.
\item \textbf{Section~\ref{supp:dataset}: Dataset Construction.} Trajectory planning, panoramic RGB-D rendering, motion annotation, and paired condition preparation.
\item \textbf{Section~\ref{supp:video_results}: Additional Panoramic Video Results.} Camera-controlled generation, backbone comparisons, and cross-chunk and loop consistency.
\item \textbf{Section~\ref{supp:recon_visuals}: Additional 3D Reconstruction Results.} Reconstructed scene layouts and held-out novel-view comparisons.
\end{itemize}
PVA and GCC denote panoramic video adaptation and geometry-guided camera control, respectively, as in the main paper.

\section{Method Details}
\label{supp:method_details}
\subsection{From an Input Panorama to Coarse Video Conditions}
\label{supp:geometry}
We estimate depth from the input RGB panorama, construct a textured mesh, and render it along the prescribed camera trajectory. The resulting coarse RGB provides appearance guidance, while projection confidence indicates its reliability and identifies regions requiring synthesis.

\setcounter{figure}{1}
\begin{figure*}[t]
\centering
\includegraphics[width=\textwidth]{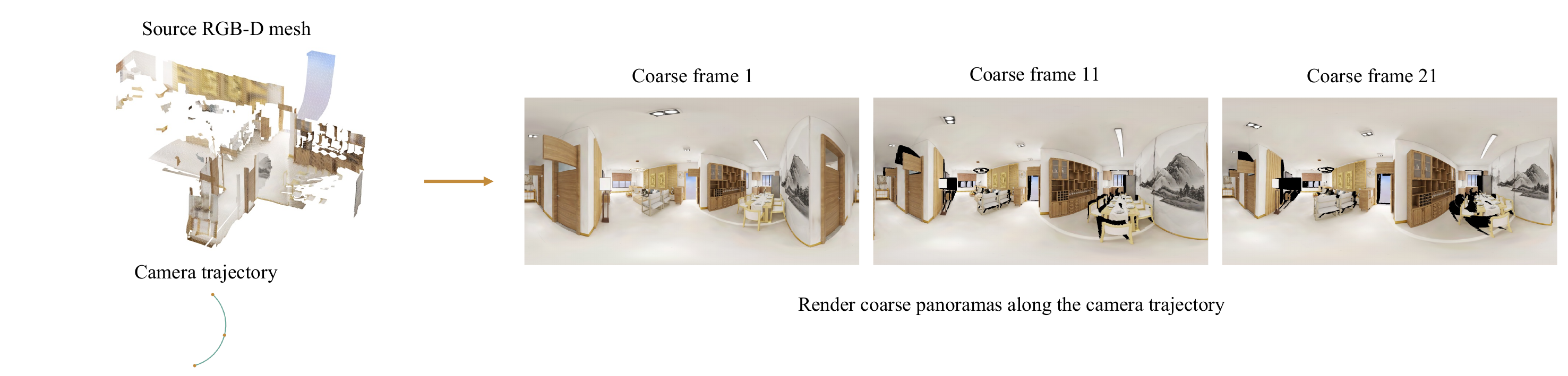}
\caption{\textbf{From source geometry to coarse panoramic video.} A textured mesh constructed from the input RGB panorama and its estimated depth is rendered along a prescribed camera trajectory to produce coarse panoramas. As the camera moves, the visible source surfaces change and previously occluded regions appear as gaps in the coarse video.}
\label{fig:supp_coarse_pipeline}
\end{figure*}
\setcounter{figure}{0}
\paragraph{Building a textured mesh from the input.}
We first estimate depth from the input panorama with MoGe-3 and convert it to radial distance $D_0$, measured from the camera center. A panorama pixel $(u,v)$ specifies a spherical ray rather than a perspective-camera ray. For image width $W$ and height $H$, its longitude and latitude are
\begin{equation}
\begin{aligned}
\lambda &= 2\pi\left(\frac{u+0.5}{W}-\frac{1}{2}\right),\\
\varphi &= \pi\left(\frac{1}{2}-\frac{v+0.5}{H}\right).
\end{aligned}
\end{equation}
The corresponding 3D point in the input-camera frame is
\begin{equation}
\begin{aligned}
\mathbf X_0(u,v) &= D_0(u,v)\,\mathbf r(\lambda,\varphi),\\
\mathbf r(\lambda,\varphi) &=
\begin{pmatrix}\cos\varphi\sin\lambda\\ \sin\varphi\\ \cos\varphi\cos\lambda\end{pmatrix}.
\end{aligned}
\end{equation}
Here $\mathbf r$ is the unit viewing ray. We use $x$-right, $y$-up, and $z$-forward coordinates, and attach the input RGB to each valid vertex. Connecting vertices from neighboring pixels forms a textured surface. Each valid four-pixel cell is split into two triangles; a cell is rejected when its depth range exceeds $0.05$ times its mean depth. This prevents connections across foreground/background boundaries. Connectivity wraps horizontally across the panorama seam. We use all source pixels (stride one), and normalize depth and camera translations by the same median valid input depth.

\begin{figure}[H]
\centering
\includegraphics[width=\linewidth]{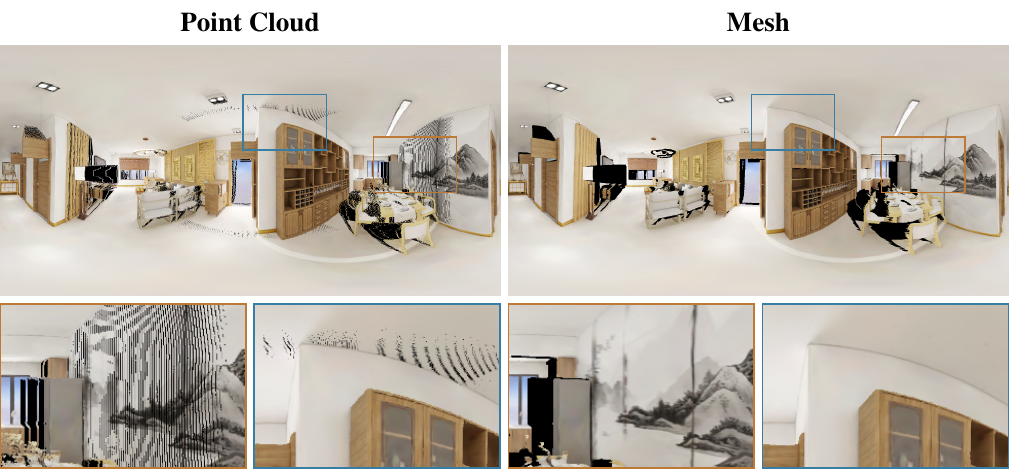}
\caption{\textbf{Surface connectivity for coarse rendering.} The same input RGB and estimated depth are rendered from a shared target camera as a point cloud (left) and a textured mesh (right). The mesh provides more continuous surface coverage in the two highlighted regions.}
\label{fig:supp_mesh_points}
\end{figure}

Figure~\ref{fig:supp_mesh_points} compares point-cloud and mesh rendering from the same estimated geometry. Point-cloud rendering can leave gaps when a viewpoint change magnifies a surface. Mesh rasterization interpolates appearance within triangles, providing more continuous coverage while depth-edge filtering prevents connections across different surfaces.

\paragraph{Rendering along the camera trajectory.}
The mesh stays fixed in the input-camera frame. For input and target camera-to-world poses $T_0$ and $T_i$, we transform its vertices by $T_i^{-1}T_0$ into target-camera coordinates. At each pose, six virtual perspective cameras cover the panorama. Depth testing selects the nearest visible surface, triangle rasterization interpolates its color, and cubemap-to-ERP sampling assembles the coarse panorama. Repeating this operation over the prescribed poses produces a temporally ordered coarse video, as illustrated in Fig.~\ref{fig:supp_coarse_pipeline}. Surfaces occluded in the input panorama are absent from the reconstructed mesh, leaving disoccluded regions unfilled in the coarse renderings.

\setcounter{figure}{2}
\paragraph{Assigning projection confidence.}
A visible surface can still carry severely stretched texture. We therefore use the main paper's confidence formulation to combine magnification, anisotropy, depth-edge risk, and occlusion-boundary risk; Fig.~\ref{fig:supp_confidence} illustrates this distinction. Here we provide the implementation settings: magnification and anisotropy tolerances are both $2$, and the four penalty weights are $1$, $0.25$, $2$, and $2$. Spherical tangent coordinates account for latitude-dependent ERP distortion, and longitude differences wrap across the panorama seam.

Depth-edge risk uses relative differences between neighboring source depths; occlusion-boundary risk uses the local visibility deficit. Both scale with camera translation relative to source distance. Unsupported pixels receive zero confidence. We align confidence with the latent spatial grid through area averaging, assigning zero confidence to invalid regions.

\begin{figure}[t]
\centering
\includegraphics[width=\linewidth]{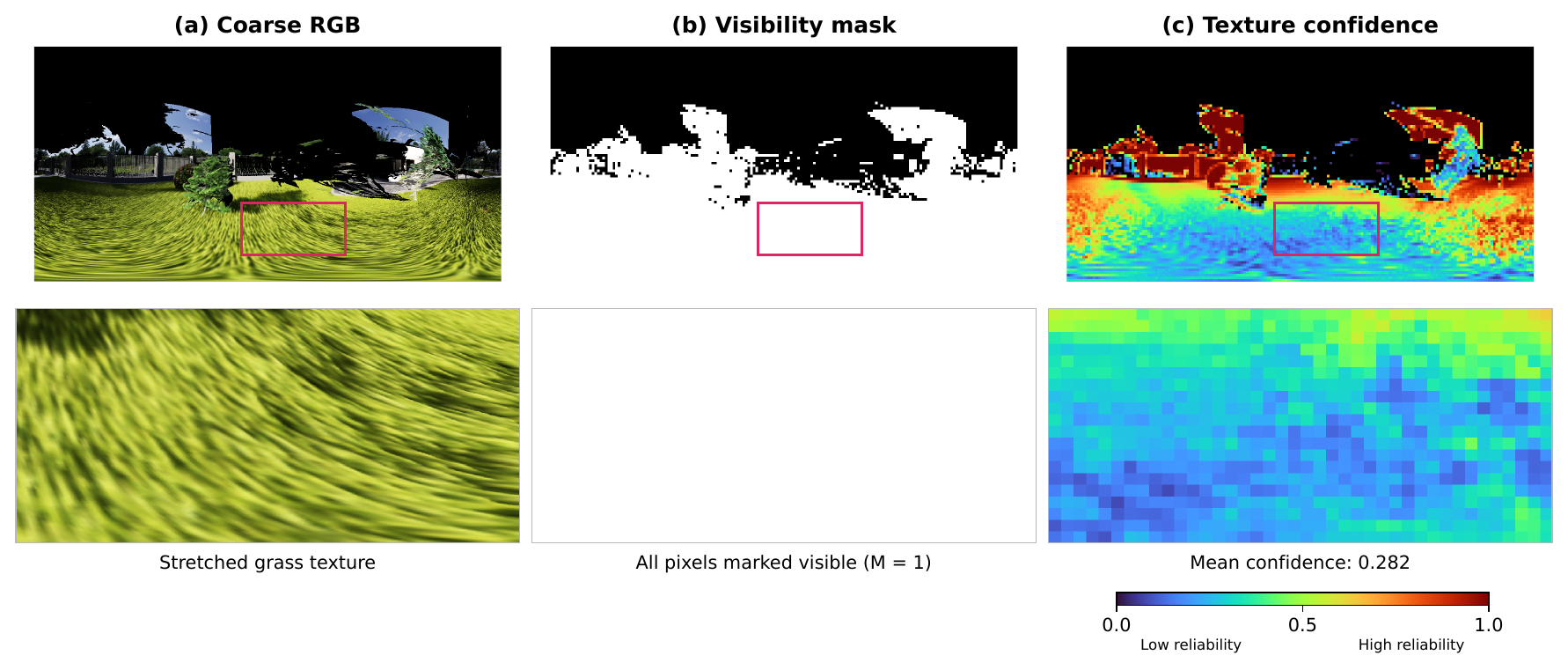}
\caption{\textbf{Visibility and projection confidence.} The grass remains visible while its projected texture is strongly stretched. Confidence, shown on a fixed $[0,1]$ scale, reduces reliance on these distorted observations even where binary visibility is valid.}
\label{fig:supp_confidence}
\end{figure}

\paragraph{Separating coverage from reliability.}
An uncovered pixel and a stretched projection require different treatment. The former contains no source observation; the latter contains observed color whose local sampling has become unreliable. Keeping coarse RGB and confidence separate preserves this distinction: GCC receives the projected structure while the confidence feature indicates how strongly its texture should be trusted. A wall can consequently retain its projected position even when its fine appearance needs regeneration. This also explains why replacing confidence with binary visibility cannot express the full conditioning signal.

\subsection{Encoding RGB and Confidence Conditions}
\label{supp:conditions}
Section~\ref{supp:geometry} constructs coarse RGB and confidence along the prescribed camera trajectory. We align these signals spatially and temporally to the video latent grid, then jointly encode their appearance and reliability as GCC condition tokens.

\paragraph{Spatial alignment.}
The frozen video VAE encodes the coarse RGB sequence using the backbone's native normalization. Confidence carries a numerical measure of projection reliability, so we preserve its meaning by area-averaging it directly to the latent spatial grid. Unsupported locations receive zero confidence. Keeping confidence as a separate feature lets the control branch distinguish projected appearance from its reliability.

\paragraph{Temporal alignment.}
Spatial correspondence alone is insufficient because the video VAE also compresses time. We therefore group confidence frames according to the backbone's temporal encoding layout and fold the frame phases into channels. Each latent time position receives four confidence phases associated with its RGB encoding. Repeated indices account for the VAE's treatment of initial frames and chunk boundaries. The RGB sequence, confidence sequence, camera poses, and target frames share the same frame selection, preserving their correspondence throughout encoding. The exact H3 and Wan index mappings are given in Sec.~\ref{supp:condition_config}.

\paragraph{Joint condition encoding.}
Once aligned, RGB latents and folded confidence are concatenated along channels and partitioned into spatial patches. A learned encoder maps each patch vector $x$ into the backbone token space:
\begin{align}
 h &= W_{\mathrm{in}}x+b_{\mathrm{in}},\\
 g &= W_2\operatorname{SiLU}\bigl(W_1\operatorname{RMSNorm}(h)+b_1\bigr)+b_2,\\
 q &= W_{\mathrm{out}}(h+g)+b_{\mathrm{out}}.
\end{align}
The residual encoder processes the concatenated RGB-latent and confidence features, then projects them to the DiT token width. Its tokens enter the control branch, whose residual outputs guide the frozen panorama-adapted backbone. Encoder dimensions are given in Sec.~\ref{supp:condition_config}.

\subsection{Maintaining and Reusing Geometric Memory}
\label{supp:cache}
A region completed in one video chunk may leave the camera view and reappear in a later chunk. 3DCache preserves that completion through four connected operations: write generated observations into memory, project historical surfaces along the new trajectory, update RGB and confidence, and guide sampling toward the recovered appearance.

\paragraph{Writing generated observations into memory.}
After a chunk is generated, its panoramas are paired with estimated radial depth and validity. As described in the main paper, estimated depths are aligned with available reference depths before constructing textured meshes. We back-project valid depth pixels into 3D vertices and connect vertices from neighboring pixels to form triangles, excluding connections across depth discontinuities. Using the source-camera pose, we transform each resulting mesh into a shared coordinate frame. 

\paragraph{Selecting and projecting historical surfaces.}
For a new trajectory, we first render the current input mesh to identify the regions it supports. We prioritize historical observations whose projected surfaces cover more of the missing regions along the target trajectory. We retain one source order for the entire trajectory, providing a consistent priority when several historical observations overlap.

The selected meshes are transformed into the target cameras and rasterized with depth testing and perspective-correct color interpolation. Periodic longitude connectivity handles the panorama seam, while depth-discontinuity and projected-size checks suppress unreliable triangles. The highest-ranked source establishes the primary historical surface. Lower-ranked sources extend it only near existing support and when their depths agree; they do not overwrite already selected historical pixels. The resulting projections carry per-pixel support weights; fusion thresholds are given in Sec.~\ref{supp:cache_config}.

\paragraph{Updating RGB and confidence.}
Historical projections fill regions outside the current input mesh while preserving its observed pixels and the initial frame. Let $F_i$ indicate these newly filled full-resolution pixels and $q_i$ their historical support weight. Their contribution to the latent-grid confidence is
\begin{equation}
 C_i'=\operatorname{clip}\left[C_i+(1-C_i)\operatorname{AreaPool}(F_iq_i),\,0,1\right].
\label{eq:supp_cache_conf}
\end{equation}
Area pooling accounts for partial coverage, so a small projected fragment does not imply full support for an entire latent cell. We re-encode the updated coarse RGB with the frozen VAE and align its confidence as in Sec.~\ref{supp:conditions}. 

\paragraph{Constructing a historical target and guiding sampling.}
The updated conditions first produce a provisional video. This video supplies context outside the historical projections, allowing a spatially partial memory to define a complete video target. We insert the projected historical texture only where the original confidence was zero and cache visibility is valid, then encode the composite with the frozen VAE to obtain $z^{\mathrm{tar}}$.

A second sampling pass uses this target to preserve the recovered appearance. For four-phase folded confidence, the guidance mask is
\begin{equation}
 M_{t,h,w}=\frac{1}{4}\sum_{c=1}^{4}C'_{c,t,h,w}\,\mathbf1[C_{c,t,h,w}=0].
\label{eq:supp_mask}
\end{equation}
The first latent time slice is set to zero, and the mask is broadcast across latent channels. Consequently, direct correction is localized to newly cache-supported regions while protecting the initial-image condition.

At flow noise level $\sigma>0$, the predicted clean latent is $\hat z=x_\sigma-\sigma v$. The clean-latent correction in the main paper can be implemented through the equivalent velocity update
\begin{align}
 v^+ &= v+\alpha M\odot\left(\frac{x_\sigma-z^{\mathrm{tar}}}{\sigma}-v\right),\\
 x_{\sigma'} &= x_\sigma+(\sigma'-\sigma)v^+.
\label{eq:supp_velocity}
\end{align}
Indeed, substitution gives $x_\sigma-\sigma v^+=\hat z+\alpha M\odot(z^{\mathrm{tar}}-\hat z)$. Unsupported locations retain the original velocity; supported locations are guided toward cached appearance. The generated observations are then stored for subsequent chunks, completing the memory update cycle.

\section{Implementation Details}
\label{supp:training}
We provide the backbone-specific settings needed to reproduce the method, organized by model configuration, training, inference, and evaluation.

\subsection{Model Configuration}
\label{supp:condition_config}
\paragraph{Backbone adaptation.}
For H3, LoRA adapts the fused query/key/value, attention output, and two feed-forward projections in all 50 DiT blocks; for Wan, it adapts the separate query, key, value, and output projections in self- and cross-attention, together with the feed-forward projections in all 40 blocks. Wan's dedicated image key/value projections are unchanged. The five control blocks are copied from H3 blocks $\{0,10,20,30,40\}$ or Wan blocks $\{0,8,16,24,32\}$, using zero-based indexing. They are initialized from the pretrained backbone before installing the PVA adapter, with zero-initialized residual output projections. Both backbones retain their native image and text conditioning.

\paragraph{Condition dimensions.}
H3 encodes 22 frames at $2880\times1440$ into 24-channel RGB latents with seven temporal positions; Wan encodes 21 frames at $1440\times720$ into 16-channel latents with six temporal positions. Both spatial grids are $90\times180$ (height $\times$ width). Four confidence phases are folded into channels, yielding 28 joint channels for H3 and 20 for Wan. Flattening $1\times2\times2$ patches gives 112 and 80 input features, respectively. The joint encoder has hidden width 512, residual MLP width 1024, and RMSNorm epsilon $10^{-6}$; its output width is 5376 for H3 and 5120 for Wan.

\paragraph{Backbone-specific temporal folding.}
The four confidence phases are stacked as channels, not averaged over time. Their grouping follows each VAE's temporal layout. For H3, the 22-frame sequence produces seven groups:
\begin{align*}
&(0,0,0,0),\ (1,2,3,4),\ (5,6,7,8),\\
&(9,10,11,12),\ (13,14,15,16),\\
&(17,17,17,17),\ (18,19,20,21).
\end{align*}
The repeated initial frame and frame 17 match H3's initial-frame encoding and chunk-local temporal reset.

Wan instead uses a $1+4k$ frame layout: the initial frame occupies one latent position, and the remaining 20 frames form five groups of four. To obtain a common 21-frame sequence from the 22-frame trajectory, we omit index 11 (the twelfth frame, with zero-based indexing) and reindex the retained frames from 0 to 20. Wan's six confidence groups are then
\begin{align*}
&(0,0,0,0),\ (1,2,3,4),\ (5,6,7,8),\\
&(9,10,11,12),\ (13,14,15,16),\ (17,18,19,20).
\end{align*}
Thus Wan has no additional reset group at index 17. Frame selection is applied jointly to RGB, confidence, poses, and targets before encoding.

\subsection{Training Configuration}
\paragraph{Optimization.}
Both stages use LoRA rank and alpha 128. PVA optimizes LoRA at $10^{-4}$; GCC freezes this adapter and optimizes the copied control blocks at $10^{-5}$ and newly introduced layers at $10^{-4}$. AdamW uses weight decay 0.01 and gradient norm clipping 1.0, with per-device batch size one and no gradient accumulation. Trainable parameters are FP32 with BF16 computation. The frozen H3 DiT is INT8, whereas the frozen Wan backbone is BF16.

\paragraph{Loss settings.}
For the PVA objective defined in the main paper, $\lambda_{\mathrm{pano}}=0.1$ and $\beta=0.5$, giving seam and edge coefficients 0.1 and 0.05. The edge band contains four latent columns on each side, and the SmoothL1 transition parameter is one. The scheduler weights the PVA loss. GCC uses only flow matching, without seam or edge losses. Wan uses flow shift 5 and empty-text dropout 0.1. Noise levels and velocity targets follow the corresponding backbone scheduler.

\subsection{Inference Configuration}
\label{supp:cache_config}
\paragraph{Video sampling.}
Standard generation uses 50 sampling steps. Wan uses flow-UniPC with shift 5, text CFG 5, and an empty negative prompt; image and geometric conditions are retained in both CFG branches. Accelerated generation uses the rank-64 LightX2V adapter for Wan or the full-critic DMAD checkpoint for INT8 H3. These acceleration components are separate from the rank-128 PVA adapter.

\paragraph{Historical surface filtering.}
For the memory selection in Sec.~\ref{supp:cache}, five uniformly spaced candidates are considered per historical chunk and three sources are retained by coverage. A triangle is rejected when its source-depth range exceeds $0.06d_{\min}$, where $d_{\min}$ is its minimum vertex depth. This relative threshold suppresses connections across depth discontinuities while adapting to scene depth. After seam unwrapping, triangles spanning more than 24 target pixels along either axis are rejected to suppress excessively stretched projections. 

\paragraph{Historical support fusion.}
To extend coverage without joining unrelated surfaces, a fallback source may fill only within four output pixels of existing support and with at most 6\% depth disagreement. Primary and fallback weights are 0.85 and 0.65, giving greater confidence to the highest-ranked surface. Coverage-normalized area averaging resizes projected colors without mixing uncovered black pixels into valid texture. A two-pixel nearest-color extension closes small target-image boundary gaps while leaving the guidance mask unchanged.

\paragraph{Cache-guided sampling.}
The accelerated Wan experiments use four steps for provisional generation and four for constrained resampling, with CFG 1, control scale 1, and guidance strength $\alpha=1$. The mask is scaled as $M\leftarrow\min(2M,1)$ to strengthen partial support while preserving zero guidance outside supported regions. All weights remain fixed; the second pass uses the historical target constructed in Sec.~\ref{supp:cache}.

\subsection{Evaluation Protocols}
\label{supp:evaluation}
\label{supp:qualitative_protocols}
\paragraph{Frame Alignment and Visual Quality Metrics.}
The main benchmark contains 200 test clips. All methods use their native output resolution, with the reference image resized to the evaluated output grid for paired metrics. Metrics use lossless frames without post-hoc registration or color correction.

When comparing 21-frame and 22-frame models at common target poses, the common 21-frame sequence excludes original index 11. The input anchor is excluded from paired frame metrics, leaving 20 non-conditioning views. Longer baseline sequences and refinement outputs are aligned using recorded time/pose correspondences.

PSNR, SSIM, and LPIPS assess reference fidelity. Frame scores are aggregated with equal clip weighting. FID and FVD use pooled feature distributions over the evaluation set. FVD uses the common video sequence, including its anchor, and the fixed I3D feature pipeline. NIQE and Q-Align provide complementary no-reference measurements. Camera adherence is evaluated separately using the pose-based protocol below.

\paragraph{Camera Control Evaluation.}
We estimate camera motion from generated RGB using VGGT. Each ERP frame is projected into four $90^\circ$ perspective views with yaw angles $0^\circ,90^\circ,180^\circ,270^\circ$ and zero pitch. The camera evaluator uses $518\times518$ perspective images. Each directional sequence is estimated separately and converted to a consistent coordinate convention before comparison.

After making both trajectories relative to their first camera, the rotation error at frame $i>0$ is
\begin{equation}
 e_R(i)=\arccos\!\left(\operatorname{clip}\left[\frac{\operatorname{tr}(\hat R_iR_i^\top)-1}{2},-1,1\right]\right).
\end{equation}
The reported $R_{\mathrm{err}}$ averages these angles in radians. For relative translations, define $s_p=\max_i\|\hat t_i\|_2$ and $s_g=\max_i\|t_i\|_2$. The normalized translation error is
\begin{equation}
 T_{\mathrm{err}}=\frac{1}{N-1}\sum_{i=1}^{N-1}\left\|\frac{\hat t_i}{s_p}-\frac{t_i}{s_g}\right\|_2.
\end{equation}
This normalization removes the global scale ambiguity of monocular pose estimation for nonzero-translation sequences. Trajectories are compared in their first-camera coordinate frames, with no additional rotation alignment. 

\paragraph{Baseline and Inference Settings.}
Matrix-3D, OmniRoam, and PanoWorld are evaluated with their respective input and trajectory conventions. Coordinate and trajectory-unit conversions preserve the common input and intended trajectory wherever supported. Imagine360 receives a ground-truth perspective video for outpainting, which contains future-view information unavailable to a single-panorama generator; it is therefore treated separately and excluded from camera-control comparison.

\paragraph{3D Reconstruction Protocol.}
For the reconstruction evaluation, each generated sequence supplies 11 fitting positions, and 10 interleaved positions are held out for evaluation. Each panorama is converted to 24 perspective directions, giving 264 fitting views and 240 evaluation views per scene. We run the same iterative 3DGS fitting code for 30,000 iterations across methods, without imposing a Gaussian-count cap. The common initialization contains 100,000 points; densification is enabled before iteration 15,000. Test renders use $720\times720$ perspective images.

For all methods, scene-scale calibration uses only generated fitting-view RGB after accounting for known baseline coordinate scaling. Neither reconstruction nor calibration uses ground-truth RGB, and calibration excludes test views and ground-truth depth.

\paragraph{Cross-chunk and loop comparisons.}
Cache ablations share the first generated chunk and use matched seeds for subsequent generation. Cross-chunk reconstruction fits ten frames per chunk with 24 perspective directions per panorama, for 30,000 iterations; paired renders use the same $360\times360$ cameras. Three $120^\circ$ arcs yield 61 sampled poses after deduplicating chunk boundaries; the start/end pose occurs twice. Each variant uses its own preceding final frame as the next input, with seeds 42, 123, and 456; the endpoint matches the initial camera pose without restoring the initial image. These experiments use the accelerated Wan sampling configuration above. For the closed-loop comparison, Matrix-3D, OmniRoam, and PanoWorld follow their official inference pipelines, generating 81 frames per chunk. The 21 prescribed key poses correspond to baseline frame indices $0,4,\ldots,80$. PanoTrek (Wan) generates 21 frames per chunk at these prescribed poses, using the accelerated sampling configuration described above. Matrix-3D uses 720p generation with 50 steps; OmniRoam uses 480p Preview and 720p Refine with 50 steps each; PanoWorld uses its released 480p model with 30 steps. OmniRoam chains the Refine endpoint. These baseline runs use no 3DCache; all methods continue from their own final frames without an endpoint reset.

\paragraph{Reconstruction displays.}
The scene cutaway is obtained by scene-specific iterative 3DGS optimization for 3,000 iterations, with scale aligned to estimated input depth; the ceiling is clipped to expose the interior. Novel-view comparisons use the common 30,000-iteration protocol above. All PanoTrek results use the coarse RGB and projection confidence conditioning described in the main paper. The panoramic baseline comparison and novel-view comparison use the H3 backbone; the backbone-detail comparison includes both Wan and H3. 

\section{Dataset Construction}
\label{supp:dataset}
We construct PanoTraj through four steps: planning camera trajectories in reconstructed environments, rendering aligned panoramic RGB-D sequences, deriving motion descriptions from camera poses, and preparing first-view geometric conditions. Each training sample binds the input panorama, target video, camera trajectory, text, coarse RGB, and confidence to the same ordered frame indices.

\begin{figure*}[t]
\centering
\includegraphics[width=\textwidth]{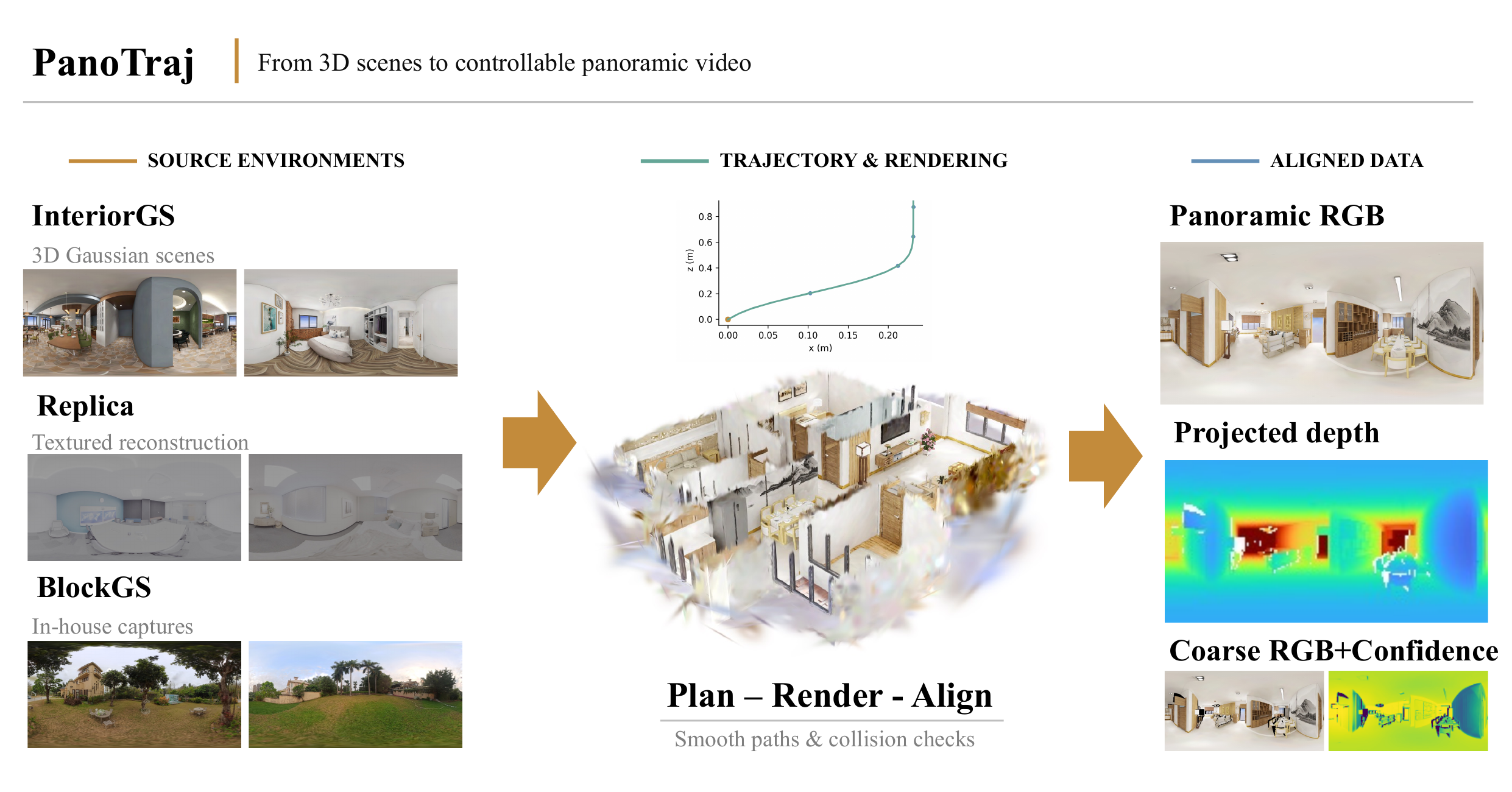}
\caption{\textbf{PanoTraj construction overview.} Source environments support camera-path sampling and panoramic rendering. RGB, radial depth, camera poses, and motion descriptions are aligned by frame; first-view reprojection supplies coarse RGB and confidence.}
\label{fig:supp_dataset_pipeline}
\end{figure*}

Figure~\ref{fig:supp_dataset_pipeline} summarizes how source geometry is converted into paired video targets and geometric conditions.

\subsection{Trajectory Planning and Clip Sampling}
A useful training trajectory must remain within a renderable region, avoid obstacles, and provide a sequence of distinct viewpoints. We therefore determine the planning domain before searching for a path, then convert the verified path into short training clips.

\paragraph{Defining the planning domain.}
For InteriorGS indoor trajectories, free pixels in an occupancy map alone do not establish a suitable rendering region: an exterior platform can be geometrically open and even appear among the room polygons. We first screen candidate regions using wall enclosure and preview RGB-D coverage. Retained room polygons define the domain, and a doorway connects regions only when both sides lie inside that domain. This prevents an exterior opening from extending an indoor trajectory onto an excluded terrace. Within this domain, occupancy, structural walls, and labeled objects determine obstacles at the camera height of 1.4\,m. Obstacle and doorway-jamb clearance then restrict where the camera can move. Region screening establishes where to plan; clearance establishes where motion is safe.

\paragraph{Connecting viewpoints into a safe path.}
Within each retained connected region, spatially distributed waypoints encourage coverage of different parts of the scene. Successive waypoints are connected by eight-connected A* search with a cost combining travel distance and low-clearance penalties. Diagonal moves cannot cut across obstacle corners. We smooth the grid path to obtain continuous camera motion and resample it along arc length. Since smoothing and resampling can introduce shortcuts, the resulting line segments are checked again against the allowed region and obstacle mask. Safe corner vertices are retained when needed; a path that still fails these checks is rejected.

\paragraph{Forming training clips.}
The verified track provides a sequence of poses for rendering. Selecting different starting positions and temporal strides produces clips with different visible content and motion magnitudes. Each canonical clip contains 22 poses; RGB, depth, and camera matrices use identical frame indices. Wan's 21-frame selection is applied afterward as specified in Sec.~\ref{supp:condition_config}. Camera orientation is stored explicitly rather than inferred from the path tangent, allowing translation and viewing direction to be controlled separately.

\subsection{Panoramic RGB-D Rendering}
Each sampled pose must provide a complete panoramic target and depth in the same coordinate system. We obtain these through virtual perspective views, assemble their colors on the spherical image grid, and convert their depths to distances along spherical rays.

\paragraph{Rendering directional observations.}
Perspective rasterizers provide the RGB and depth needed for panoramic supervision. For Gaussian scenes, we place six virtual pinhole cameras at the sampled optical center, facing forward, backward, left, right, up, and down, each with a $90^\circ$ field of view. Composing these six rotations with the trajectory pose keeps all views at the same position and time. Rendering $1440\times1440$ images gives a cubemap covering the full sphere, from which we construct a $2880\times1440$ ERP panorama.

\paragraph{Mapping perspective views to the panorama.}
For each ERP pixel $(u,v)$, its longitude and latitude define a unit viewing ray. We convert the $y$-up ray in Sec.~\ref{supp:geometry} to the renderer's $y$-down convention using $y=-\sin\varphi$, keeping $x$ and $z$ unchanged. Face-camera $y$ and image row coordinates both increase downward, hence the positive sign in the vertical projection below. We select the cubemap face whose forward axis is most aligned with this ray, then express the ray in that face's camera coordinates as $\mathbf d=(d_x,d_y,d_z)^\top$, with $d_z>0$. Its corresponding face-image location is
\begin{equation}
 u_f=f_x\frac{d_x}{d_z}+c_x,\qquad
 v_f=f_y\frac{d_y}{d_z}+c_y.
\end{equation}
Here $(u_f,v_f)$ are continuous pixel coordinates in the selected perspective image, $f_x,f_y$ are focal lengths in pixels, and $(c_x,c_y)$ is its principal point. For our face resolution and field of view, $f_x=f_y=c_x=c_y=720$. Bilinear sampling at $(u_f,v_f)$ transfers RGB, opacity, and camera-axis depth to the ERP pixel. Thus each panoramic pixel receives an observation from the face aligned with its viewing direction.

\paragraph{Obtaining geometrically consistent depth.}
The face renderer reports depth along its optical axis, whereas spherical back-projection requires distance along the viewing ray. Let $Z_f(u_f,v_f)$ be the sampled camera-axis depth in the selected face. Because $\mathbf d$ is a unit vector, the radial depth at panoramic pixel $(u,v)$ is
\begin{equation}
 D(u,v)=\frac{Z_f(u_f,v_f)}{d_z}.
\end{equation}
This conversion makes depths from different cubemap directions consistent for subsequent mesh construction. Low-opacity pixels and nonfinite or out-of-range depths are marked invalid. We store each panorama with its radial depth, validity, and camera-to-world pose, giving the next stages aligned appearance and geometry.

\paragraph{Preserving depth at face transitions.}
The optical-axis depth of a surface changes with the orientation of the virtual camera, even when the surface point and optical center are fixed. Using those values directly as panorama depth would make reconstructed distances depend on the selected cubemap face. Dividing by $d_z$ removes this directional dependence before back-projection. RGB, radial depth, and validity can then describe the same surface point across face transitions, which is necessary when those observations are later connected into a mesh.

\subsection{Motion Annotation and Text Conditioning}
The known camera trajectory provides motion labels without estimating movement from rendered images. We express poses relative to the initial camera and normalize translation by the first-frame scene scale, making displacement descriptions comparable across scenes.

\paragraph{Describing translation and orientation.}
Maximum excursion from the starting position determines motion magnitude. Comparing endpoint displacement with excursion distinguishes returning paths, while comparison with total path length distinguishes curved and mostly straight paths. Dominant translation components determine left/right, up/down, and forward/backward directions in the initial camera frame. Relative rotations supply a separate turning or tilting description; small components are suppressed to avoid labeling numerical jitter as motion. Separating translation from orientation distinguishes moving sideways from rotating in place.

\paragraph{Constructing the text condition.}
We combine this pose-derived motion description with a reviewed spatial category obtained from representative first-frame imagery. The category describes the environment, and the motion description specifies how the camera explores it. Their combination is formatted and encoded for the corresponding backbone after alignment with the sampled clip.

\setcounter{dbltopnumber}{1}
\renewcommand{\dbltopfraction}{0.9}
\makeatletter
\setlength{\@dblfptop}{0pt}
\makeatother
\setcounter{figure}{4}
\begin{figure*}[t]
\centering
\includegraphics[width=\textwidth]{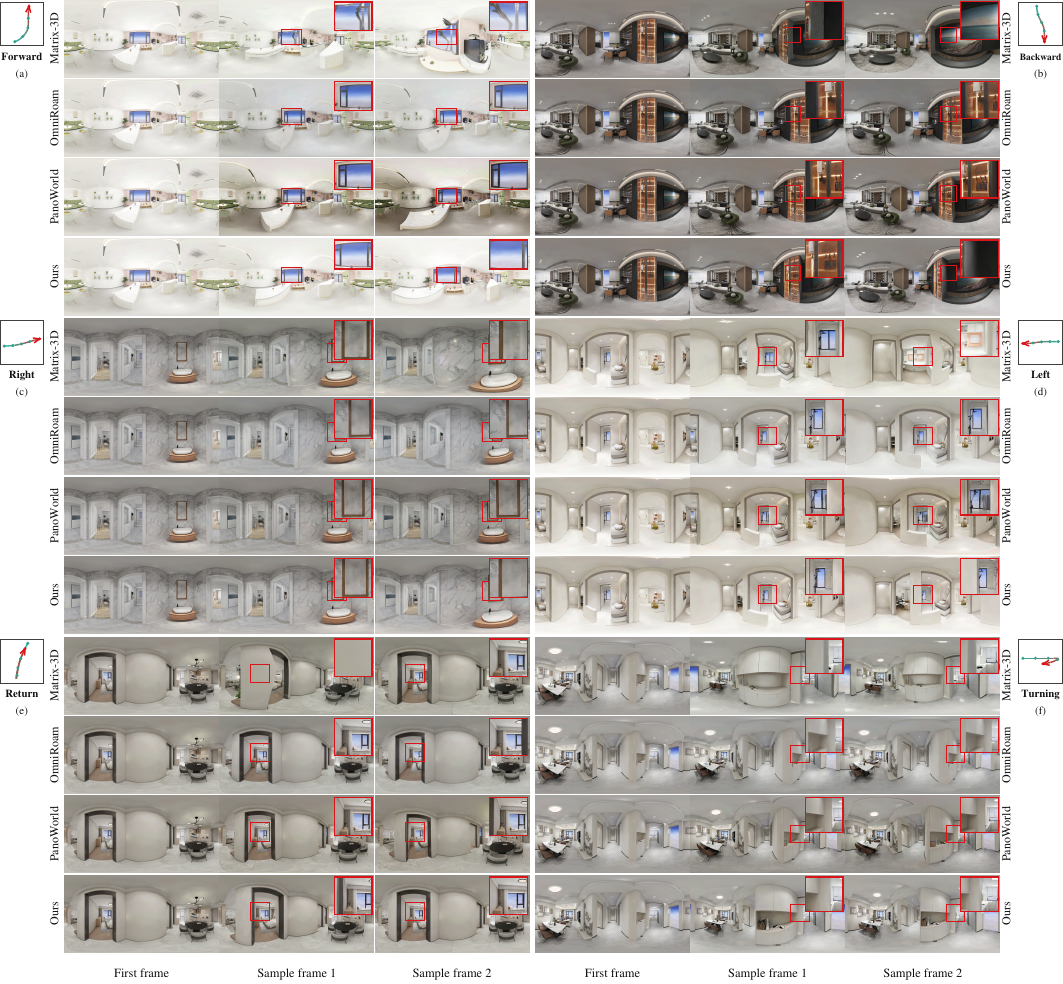}
\caption{\textbf{Panoramic video generation.} Each group compares Matrix-3D, OmniRoam Refine, PanoWorld, and PanoTrek (H3) at the input and two subsequent views. The examples test appearance preservation under translation, turning, and near-return motion.}
\label{fig:supp_six_trajectories}
\end{figure*}

\subsection{Geometric Condition Preparation}
The final step pairs complete target observations with the partial geometric guidance available from the input view. Target panoramas are rendered from the full source environment. In contrast, coarse RGB and confidence are constructed only from the clip's first RGB-D observation and prescribed poses, using the mesh projection procedure in Sec.~\ref{supp:geometry}. A surface newly revealed behind an occluder is therefore present in the target but absent from the coarse condition. This pairing teaches the generator to complete unseen content while following the supplied geometry.

We encode target and coarse videos with the backbone's frozen VAE and align confidence using Sec.~\ref{supp:conditions}. RGB, depth, poses, text, and conditions retain a common sample identity and frame ordering. H3 and Wan use their respective encodings, while all clips and encoded versions of a scene share the same train/test assignment.

\setcounter{figure}{5}
\begin{figure*}[t]
\centering
\includegraphics[width=\textwidth]{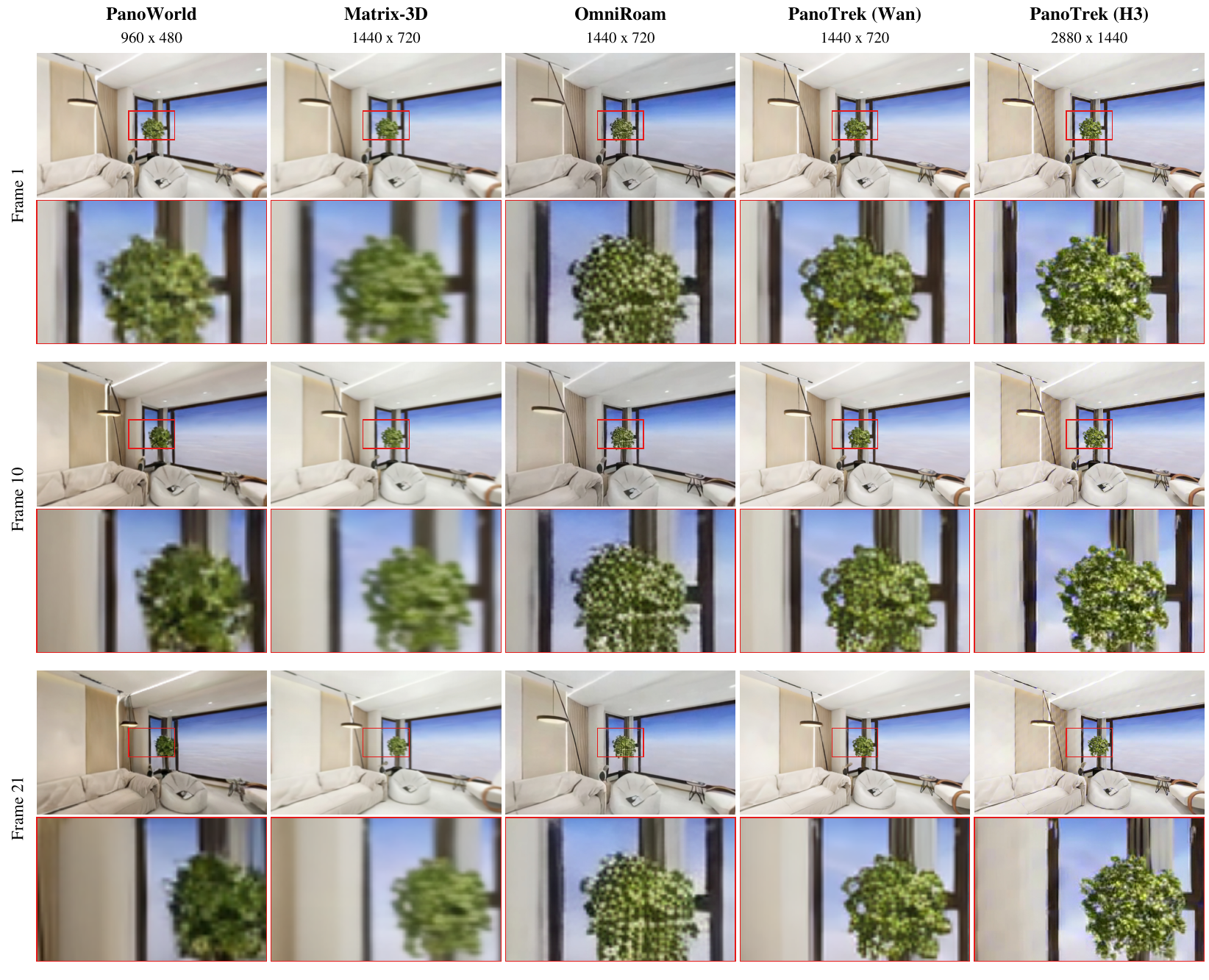}
\caption{\textbf{Visual detail across methods and native resolutions.} PanoWorld, Matrix-3D, OmniRoam Refine, and the Wan/H3 variants of PanoTrek are ordered by native output resolution. The high-resolution PanoTrek (H3) output preserves clearer local detail across the sampled views.}
\label{fig:supp_native_resolution_visual}
\end{figure*}

\setcounter{figure}{6}
\begin{figure*}[t]
\centering
\includegraphics[width=\textwidth]{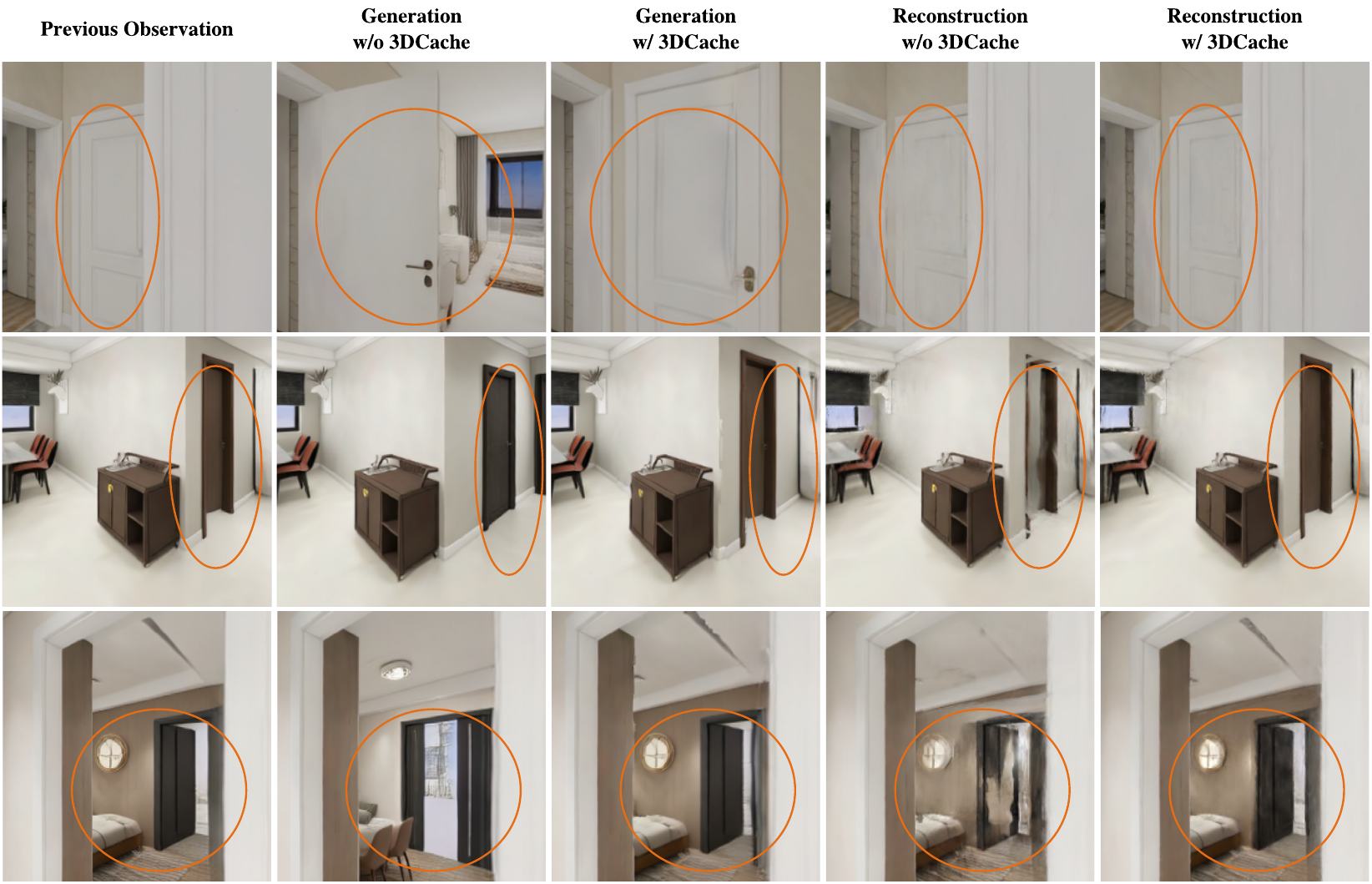}
\caption{\textbf{Effect of 3DCache on reconstructed geometry.} Repeated generation without memory changes previously completed doors and room contents, producing conflicting observations for 3DGS fitting. Reusing historical surfaces with 3DCache improves agreement between chunks and preserves more coherent reconstructed boundaries.}
\label{fig:supp_cache_reconstruction}
\label{fig:supp_twochunk_seedchange}
\end{figure*}

\setcounter{figure}{7}
\begin{figure*}[t]
\centering
\includegraphics[width=\textwidth]{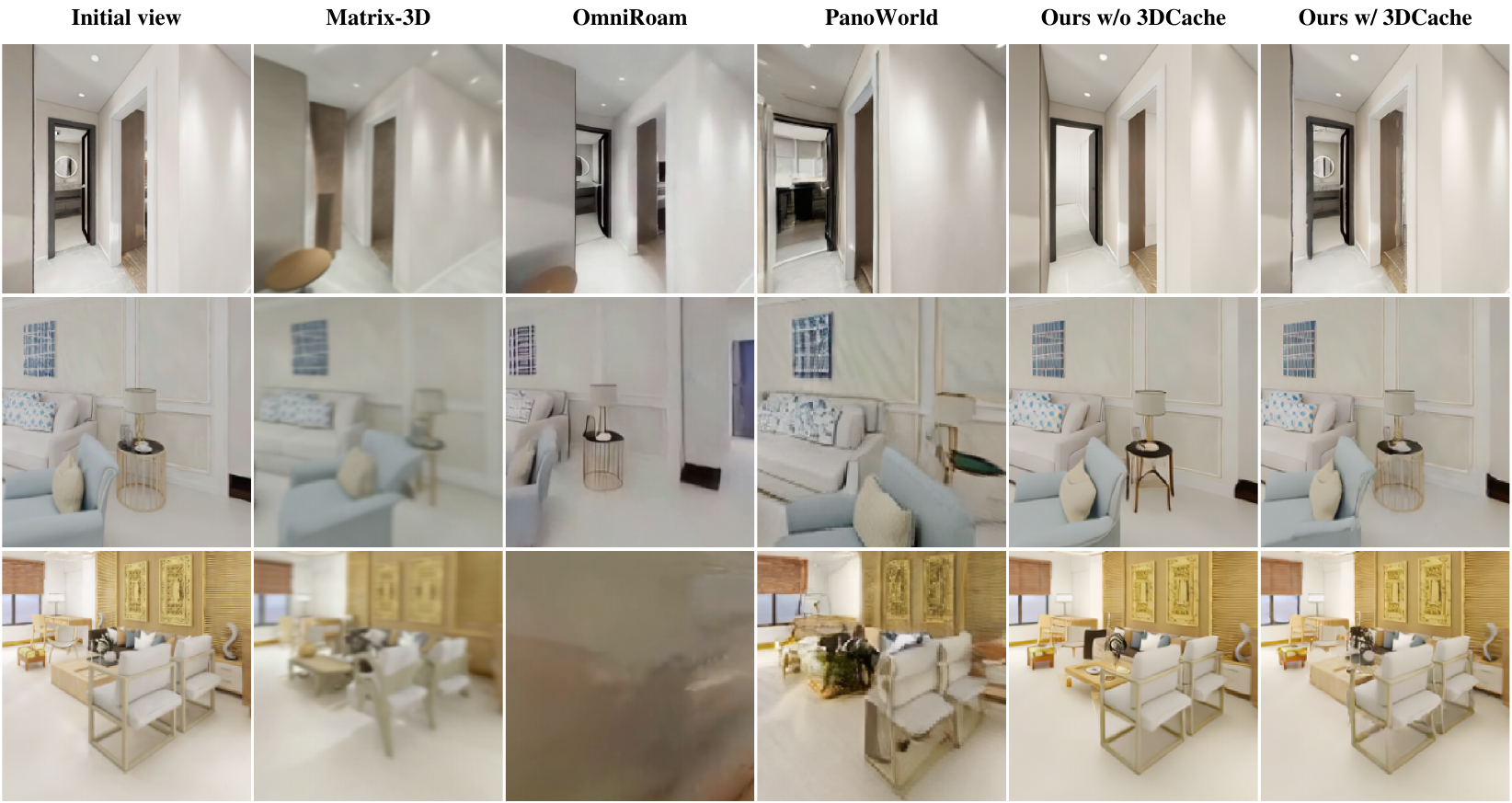}
\caption{\textbf{Loop consistency across methods.} Initial views and final outputs after following the same prescribed closed-loop trajectory. Columns compare Matrix-3D, OmniRoam Refine, PanoWorld, and PanoTrek without and with 3DCache. Each method continues from its own preceding final frame. The return views test whether the initial room layout and object identities survive the intervening exploration.}
\label{fig:supp_loop_consistency}
\end{figure*}

\section{Additional Panoramic Video Results}
\label{supp:video_results}
We examine appearance under camera motion, visual detail across backbones, and consistency when previously generated regions are revisited. Experimental configurations are provided in Sec.~\ref{supp:qualitative_protocols}.

\subsection{Panoramic Video Generation}
\label{supp:trajectory_comparison}
\paragraph{Appearance under camera motion.}
Figure~\ref{fig:supp_six_trajectories} compares methods under forward, backward, lateral, turning, and near-return trajectories. These motions require the generator to change foreground--background relationships while retaining recognizable scene content. The highlighted windows, framed wall features, and shelving provide local references for examining this behavior. In the displayed baseline results, some of these structures lose definition or change appearance as the viewpoint moves. PanoTrek preserves more coherent boundaries and finer appearance details across the sampled views, while the surrounding scene changes with the camera trajectory.

The coarse renderings move observed surfaces according to the prescribed cameras, anchoring synthesis to their projected locations. PanoTrek can therefore refine texture and complete disoccluded regions while retaining the surrounding layout, rather than regenerating that layout independently at each viewpoint.

\subsection{Visual Detail Across Methods and Resolutions}
\label{supp:native_resolution_visual}
\paragraph{Resolution and detail preservation.}
The preceding comparison examines room structure under camera motion. Figure~\ref{fig:supp_native_resolution_visual} focuses on the finer appearance retained within that structure, with outputs ordered from $960\times480$ to $2880\times1440$. A panorama allocates pixels over the full sphere, so any particular viewing direction receives only a fraction of the total image resolution. Increasing panoramic resolution provides more samples for the same angular region and allows smaller texture variations and object boundaries to be represented.

In the displayed examples, the lower-resolution outputs reproduce the overall room appearance, while the high-resolution PanoTrek (H3) output retains clearer local texture in the plant region. This difference remains visible at successive output times: the additional detail is maintained during camera motion, rather than appearing only in the conditioning image. The result illustrates the benefit of higher-resolution generation for exploration, where users inspect local perspective views rather than the entire panorama at once.

The comparison therefore connects output resolution to the detail available during viewing: similar room-level appearance can conceal substantial differences in local clarity. PanoTrek's high-resolution configuration preserves more of this detail across the generated sequence. Since the columns also differ in generator and backbone, the figure compares complete systems at their native resolutions rather than measuring the effect of resolution alone.

\subsection{Effect of 3DCache}
\label{supp:twochunk_seedchange}
\paragraph{Repeated completion of previously observed regions.}
High local image quality does not ensure that successive chunks describe the same environment. As a trajectory revisits a region, content completed in an earlier chunk may need to be generated again, even though it is no longer visible in the current reference frame. Figure~\ref{fig:supp_cache_reconstruction} examines this situation. Without 3DCache, the previously closed door becomes open, the doorway and adjacent wall change configuration, and the bedroom opening reveals different contents. The later completion can look plausible by itself while contradicting the earlier observation.

\paragraph{Preserving the earlier completion.}
3DCache addresses this ambiguity by retaining generated surfaces and projecting them along the new trajectory. The previous completion then supplies spatially aligned guidance where the current reference image provides no observation. In the paired results, the door state, opening boundaries, and room contents more closely match the earlier chunk. The benefit is therefore preservation of a particular established scene, rather than merely producing another plausible view of a similar room.

\paragraph{Consequences for reconstruction.}
These differences become consequential when the chunks are used together for iterative 3DGS fitting. Conflicting observations ask one static representation to explain different door states or room layouts at the same location. In the reconstructions without 3DCache, the affected regions exhibit blurred or disrupted surfaces. With historical guidance, the generated observations agree more closely, and the fitted doorways and room boundaries remain more coherent. The comparison links the three stages directly: revisiting a region creates a repeated-completion ambiguity, geometric memory constrains that completion, and the resulting agreement improves the observations available for reconstruction.

\subsection{Loop Consistency}
\label{supp:loop_consistency}
\paragraph{Consistency after extended exploration.}
The previous experiment examines agreement in revisited regions across chunks. A closed-loop trajectory extends this test to the scene recovered after a sequence of camera movements. In Fig.~\ref{fig:supp_loop_consistency}, each method starts from the same panorama and follows the prescribed loop. Every subsequent chunk uses that method's own preceding final frame as its reference; the original image is not restored at the endpoint. The final view must therefore recover the initial scene from the generated history.

\paragraph{Error propagation across chunks.}
This chaining also passes errors from one chunk to the next. If the preceding endpoint has lost detail or changed the scene layout, the next chunk receives that degraded observation as its reference. Repeated conditioning on such outputs can accumulate appearance and geometric errors, contributing to blur and structural drift in later chunks. This provides a possible explanation for the softer final views in the baseline comparison, alongside the differences in each method's native generation quality.

Matrix-3D produces softer return views and altered object boundaries in the displayed scenes. OmniRoam retains recognizable structures in the first two cases, but their placement shifts. In the third case, its generated views deviate from the prescribed camera positions. The resulting scene layout intersects the subsequent camera path, causing severe degradation as generation continues. PanoWorld also changes the doorway appearance and furniture arrangement. These return views show that continuity at a chunk boundary is insufficient to preserve the environment throughout a longer trajectory.

\paragraph{Recovering content beyond the boundary frame.}
The PanoTrek ablation isolates the role of historical guidance in this setting. Without 3DCache, the bathroom mirror and vanity disappear, and the side table's cylindrical support changes into a different leg structure. With 3DCache, these elements and the living-room table arrangement are better retained. A shared boundary frame carries the immediately visible context forward, whereas geometric memory also provides observations of regions that have left the view. Reprojecting those observations gives later chunks a reference to earlier scene content rather than relying solely on the latest generated image.

\setcounter{figure}{9}
\begin{figure*}[t]
\centering
\includegraphics[width=\textwidth]{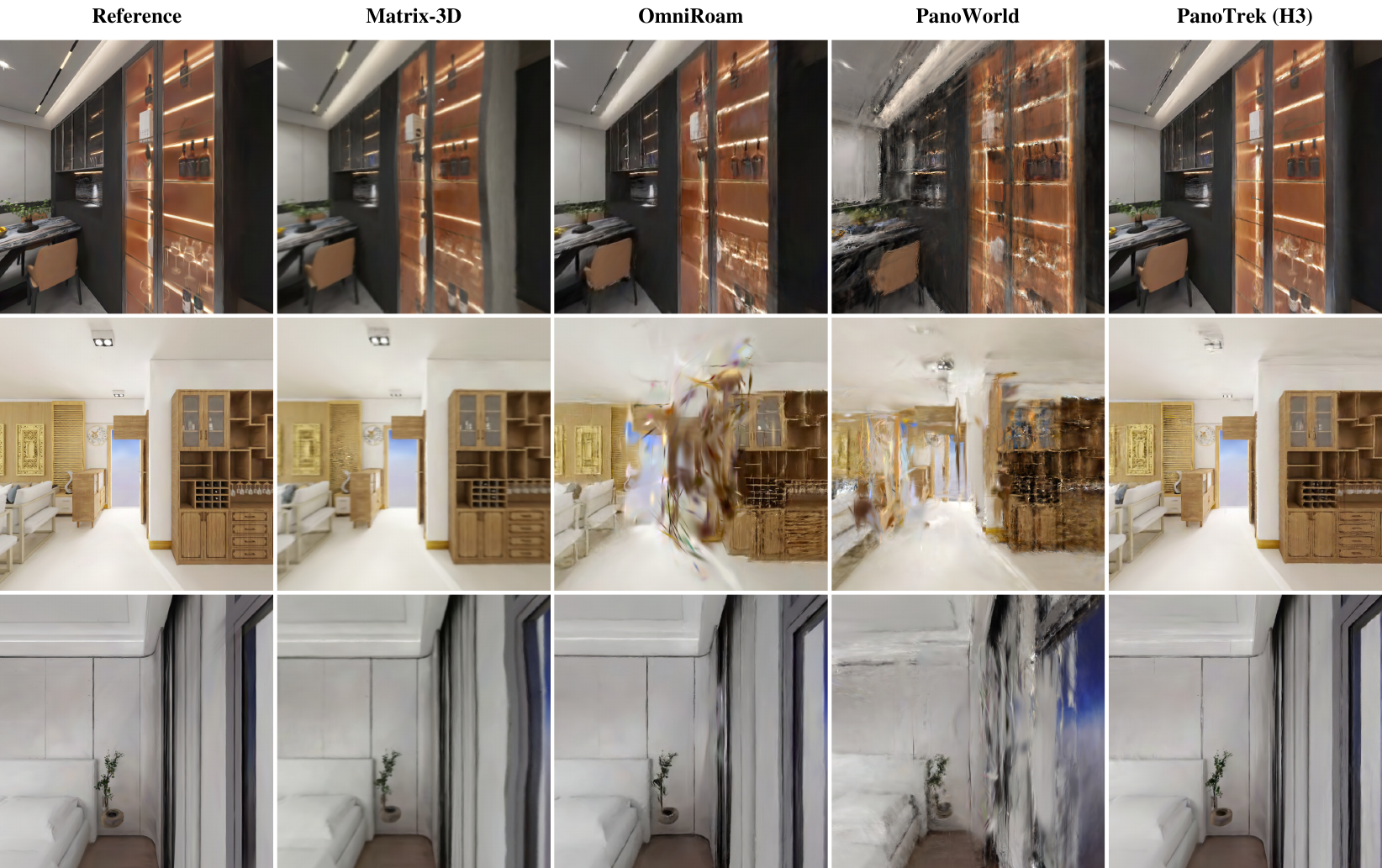}
\caption{\textbf{Qualitative comparison of novel-view synthesis.} Reference images and held-out views rendered from 3DGS reconstructions of Matrix-3D, OmniRoam, PanoWorld, and PanoTrek (H3) videos. All methods use the same fitting budget and matched evaluation cameras.}
\label{fig:supp_reconstruction_examples}
\end{figure*}

\section{Additional 3D Reconstruction Results}
\label{supp:recon_visuals}
We first reconstruct a shared 3D scene from generated panoramas, then compare novel-view renderings across methods.

\subsection{Reconstruction from Generated Panoramas}
\label{supp:reconstruction_showcase}
\paragraph{From panoramic video to 3DGS.}
The generated panoramas provide observations from successive camera positions. We convert them into overlapping perspective views, with each view inheriting the panorama's camera center and a known viewing rotation. For the scene in Fig.~\ref{fig:supp_reconstruction_showcase}, we fit a shared 3D Gaussian representation to these observations through scene-specific iterative optimization, using the prescribed camera poses. Section~\ref{supp:evaluation} gives the reconstruction settings.

\paragraph{Angular coverage and translational evidence.}
Perspective views extracted from one panorama share an optical center. They cover different directions and expose surfaces throughout the room, but they do not introduce a new translational baseline. That baseline comes from camera movement between generated panoramas. A furniture edge shifts relative to the wall behind it as the camera moves, while a doorway reveals surfaces previously hidden by the adjacent wall. Fitting these observations in a shared coordinate frame constrains the relative placement of foreground and background surfaces. Panoramic coverage and camera translation thus play different roles: the former supplies observations around each camera position, and the latter relates scene structure across positions. These views are fitted jointly to one set of Gaussians. Agreement between overlapping observations allows the same surfaces to explain the sequence, while newly exposed regions extend the reconstructed scene.

\setcounter{figure}{8}
\begin{figure}[H]
\centering
\includegraphics[width=\linewidth]{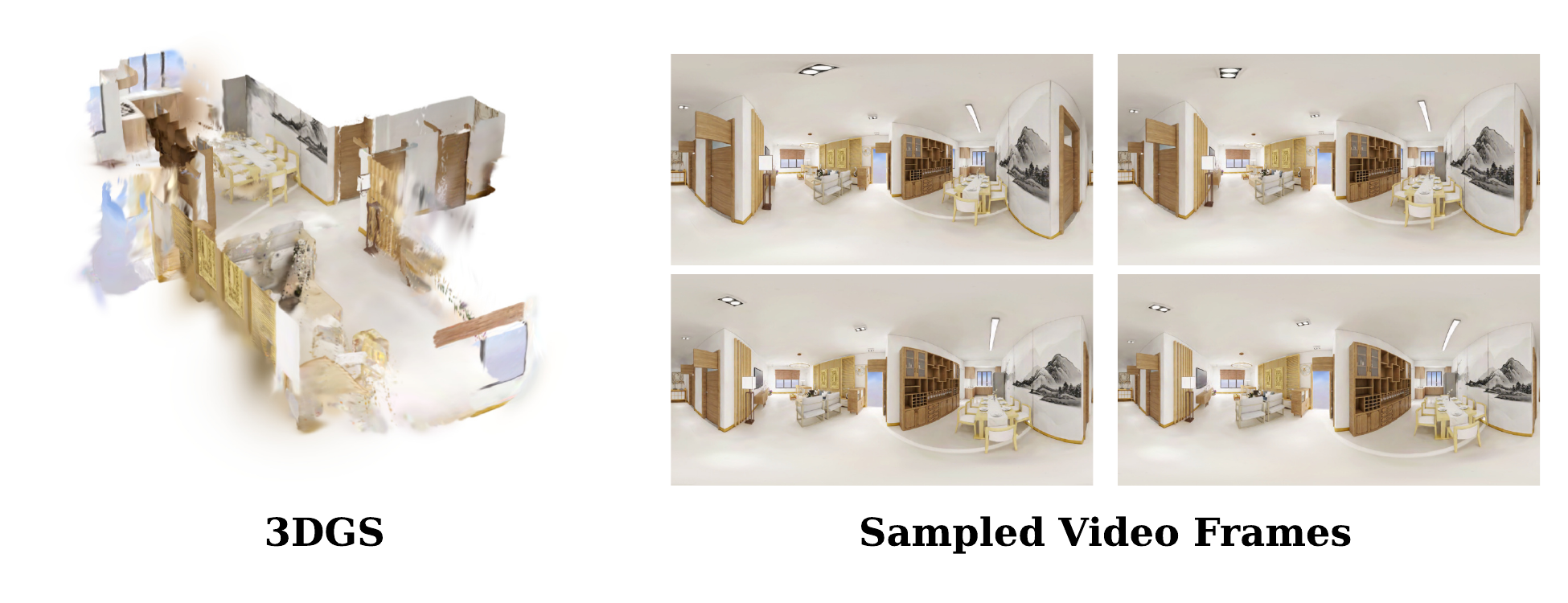}
\caption{\textbf{3D reconstruction from generated panoramas.} Left: the scene reconstructed through iterative 3DGS optimization. Right: sampled panoramic frames used for fitting. The ceiling and outermost splats are clipped to reveal the interior layout.}
\label{fig:supp_reconstruction_showcase}

\end{figure}
\setcounter{figure}{10}

\paragraph{Recovered scene layout.}
The cutaway in Fig.~\ref{fig:supp_reconstruction_showcase} reveals dining furniture, shelving, and openings within the same reconstructed room. Their arrangement can be inspected from outside the panoramic camera path, rather than inferred separately from each video frame. In particular, furniture and surrounding walls occupy distinct parts of the reconstruction instead of being represented only as texture in a single spherical image. The generated sequence therefore supplies observations that can be consolidated into a scene with recoverable interior structure.

\paragraph{Observation consistency and reconstruction quality.}
The recovered layout depends on agreement between the observations used for fitting. The door example in Fig.~\ref{fig:supp_cache_reconstruction} makes this requirement concrete: a closed-door image places a surface across the opening, whereas an open-door image reveals the space behind it. Those observations impose conflicting constraints on a static reconstruction. The resulting error concerns both the door boundary and the visibility of the room behind it, so increasing texture sharpness alone cannot resolve the disagreement.

With 3DCache, later views retain more of the earlier door state and background content. This gives the optimizer observations with a more consistent foreground boundary and occlusion relationship. The more coherent reconstruction thus follows from agreement in the generated scene structure, not simply from the visual quality of individual frames. Together, the room cutaway and the paired cache reconstructions show how camera-controlled generation supplies scene coverage and how historical consistency makes that coverage useful for fitting a shared 3D representation.

\subsection{Novel-View Synthesis Comparison}
\paragraph{Held-out viewpoints.}
Figure~\ref{fig:supp_reconstruction_examples} evaluates the reconstructed scenes from camera positions excluded from fitting. All methods use the same reconstruction budget and matched evaluation cameras. In the shelving example, the PanoTrek reconstruction retains the cabinet divisions and illuminated horizontal structures. In the dining area, it preserves the central opening and the shelving layout, whereas some baseline reconstructions exhibit smeared or fragmented surfaces. In the bedroom, wall-panel boundaries and the curtain edge remain more clearly defined. The retained cabinet divisions and wall edges show that detail survives the conversion from generated images to a shared representation.

Interleaved test cameras evaluate consistency between fitting positions, which sharp fitting images alone cannot establish. PanoTrek's clearer shelving and wall boundaries in held-out views indicate more coherent interpolation. Together with the cutaway, these results show the recovered layout and novel-view appearance.
\end{document}